\documentclass{article}

\usepackage{amsmath,amsthm,amssymb}

\newcommand\cut[1]{}

\newcommand{\seqt}{{\leq t}}
\newcommand{\seqT}{{\leq T}}
\newcommand{\st}{{<t}}

\newcommand{\squishlist}{
   \begin{list}{$\bullet$}
    { \setlength{\itemsep}{0pt}      \setlength{\parsep}{3pt}
      \setlength{\topsep}{3pt}       \setlength{\partopsep}{0pt}
      \setlength{\leftmargin}{1.5em} \setlength{\labelwidth}{1em}
      \setlength{\labelsep}{0.5em} } }

\newcommand{\squishlisttwo}{
   \begin{list}{$\bullet$}
    { \setlength{\itemsep}{0pt}    \setlength{\parsep}{0pt}
      \setlength{\topsep}{0pt}     \setlength{\partopsep}{0pt}
      \setlength{\leftmargin}{2em} \setlength{\labelwidth}{1.5em}
      \setlength{\labelsep}{0.5em} } }

\newcommand{\squishend}{
    \end{list}  }

\newcommand{\myvec}[1]{\mathbf{#1}}
\newcommand{\myvecsym}[1]{\boldsymbol{#1}}

\newcommand{\vphi}{\myvecsym{\phi}}

\newcommand{\vPsi}{\myvecsym{\Psi}}
\newcommand{\vtheta}{\myvecsym{\theta}}

\newcommand{\vc}{\myvec{c}}

\newcommand{\vg}{\myvec{g}}
\newcommand{\vh}{\myvec{h}}

\newcommand{\vw}{\myvec{w}}

\newcommand{\vx}{\myvec{x}}

\newcommand{\vz}{\myvec{z}}

\newcommand{\vM}{\myvec{M}}

\newcommand{\E}{\mathbb{E}}

\newcommand{\be}{\begin{equation}}
\newcommand{\ee}{\end{equation}}
\newcommand{\bea}{\begin{eqnarray}}
\newcommand{\eea}{\end{eqnarray}}
\newcommand{\beaa}{\begin{eqnarray*}}
\newcommand{\eeaa}{\end{eqnarray*}}

\DeclareMathAlphabet{\mathpzc}{OT1}{pzc}{m}{n}

\usepackage{iclr2017_conference, times}
\usepackage[]{amsmath}
\usepackage{natbib}
\usepackage{graphicx}
\usepackage{authblk}
\usepackage{enumitem}
\usepackage{lscape}
\graphicspath{{figures/}}
\usepackage{algorithm}
\usepackage[noend]{algpseudocode}
\usepackage{caption}
\usepackage{subcaption}

\usepackage[export]{adjustbox}
\usepackage{enumitem}
\usepackage{rotating}

\title{\begin{center}Generative Temporal Models with Memory\end{center}}
\author[$$]{\textbf{Mevlana Gemici$^*$, Chia-Chun Hung$^*$, Adam Santoro$^*$, Greg Wayne$^*$}\\
\textbf{Shakir Mohamed, Danilo J. Rezende, David Amos, Timothy Lillicrap}}
\affil[$$]{DeepMind, London}

\begin{document}
\maketitle
\let\thefootnote\relax\footnotetext{$^*$Equal Contributions.}
\begin{abstract}
We consider the general problem of modeling temporal data with long-range dependencies, wherein new observations are fully or partially predictable based on temporally-distant, past observations. A sufficiently powerful temporal model should separate predictable elements of the sequence from unpredictable elements, express uncertainty about those unpredictable elements, and rapidly identify novel elements that may help to predict the future. To create such models, we introduce \emph{Generative Temporal Models} augmented with external memory systems. They are developed within the variational inference framework, which provides both a practical training methodology and methods to gain insight into the models' operation. We show, on a range of problems with sparse, long-term temporal dependencies, that these models store information from early in a sequence, and reuse this stored information efficiently. This allows them to perform substantially better than existing models based on well-known recurrent neural networks, like LSTMs.
\end{abstract}

\section{Introduction}

Many of the data sets we use in machine learning applications are sequential, whether these be natural language and speech processing data, streams of high-definition video, longitudinal time-series from medical diagnostics, or spatio-temporal data in climate forecasting. 
Generative Temporal Models (GTMs) are a core requirement for these applications. Generative Temporal Models are also important components of intelligent agents, as they permit counterfactual reasoning, physical predictions, robot localisation, and simulation-based planning among other capacities \citep{sutton1991dyna, deisenroth2011pilco, watter2015embed, levine2014learning, assael2015data}. 
These tasks require models of high-dimensional observation sequences and contain complex, long temporal dependencies---requirements that most available GTMs are unable to fulfil. Developing such GTMs is the aim of this paper.


Many GTMs---whether they are linear or nonlinear, deterministic or stochastic---assume that the underlying temporal dynamics is governed by low-order Markov transitions and use fixed-dimensional sufficient statistics. Examples of such models include Hidden Markov Models \citep{rabiner1989tutorial}, and linear dynamical systems such as Kalman filters and their non-linear extensions \citep{kalman1960new, ghahramani1996parameter, krishnan2015deep}. The fixed-order Markov assumption used in these models is insufficient for characterising many systems of practical relevance. \citet{bialek2001predictability} quantitatively show that Markov assumptions fail to describe physical systems with long-range correlations, and fail to approximate the long-distance dependencies in written literature.
Models that instead maintain information in large, variable-order histories, e.g., recurrent neural networks \citep{pearlmutter1995gradient}, can have significant advantages over ones constrained by fixed-order Markov assumptions.

Most recently proposed GTMs, like variational recurrent neural networks (VRNNs) \citep{chung2015recurrent} and Deep Kalman Filters \citep{krishnan2015deep}, are built upon well-known recurrent neural networks, like Long Short-Term Memory (LSTM) \citep{hochreiter1997long} and Gated Recurrent Units (GRUs) \citep{chung2015gated}. In principle, these recurrent networks can solve variable-order Markovian problems, as the additive dynamics are designed to store and protect information over long intervals. In practice, they scale poorly when higher capacity storage is required. These RNNs are typically densely connected, so the parametric complexity of the model can grow quadratically with the memory capacity.
Furthermore, their recurrent dynamics must serve two competing roles: they must preserve information in a stable state for later retrieval, and they must perform relevant computations to distill information for immediate use. These limitations point to the need for RNNs that separate memory storage from computation. 

Recurrent networks that successfully separate memory storage from computation have been developed for several settings such as algorithm learning \citep{graves2014neural,grefenstette2015learning,joulin2015inferring,reed2015neural,riedel2016programming,vinyals2015pointer}, symbolic reasoning \citep{weston2014memory,sukhbaatar2015end}, and natural language processing \citep{bahdanau2014neural,kumar2015ask,hermann2015teaching,kadlec2016text}. 
These recurrent networks store information in a memory buffer and use differentiable addressing mechanisms (often called ``differentiable attention'') to efficiently optimise reading from and writing to memory. The particular details of a system's memory access mechanisms play a critical role in determining its data efficiency.

We demonstrate that generative temporal models \emph{with memory} (GTMMs) exhibit a significantly enhanced capacity to solve tasks involving complex, long-term temporal dependencies. We develop a common architecture for generative temporal models and study four instantiations that each use a different type of memory system. These four models allow us to show how different memory systems are adapted to different types of sequential structure and the resulting impact on modelling success, data-efficiency, and generation quality. Our models are distinct from the one presented by \citet{li2016learning}, who developed a deep generative model for images posessing an attentional lookup mechanism. For \citet{li2016learning}, the memory contains a table of parameters that is not updated within a sequence. Instead, it is a table of biases that is jointly optimised for end-to-end performance. 
In contrast, our systems dynamically update the memory within each sequence. 

We structure our discussion by first describing the general approach for designing generative temporal models and performing variational inference (Section 2). We then compare GTMMs with VRNNs \citep{chung2015recurrent} on a set of visual sequence tasks designed to stress different problems that arise when modelling information with long time dependencies. Finally, we make strides toward scaling the models to richer perceptual modelling in a three-dimensional environment. In the process, we make the following technical contributions:
\begin{itemize}[noitemsep,topsep=0pt,parsep=0pt,partopsep=0pt, leftmargin=*]
\item We develop a general architecture for generative models with memory. This architecture allows us to develop GTMMs based on four memory systems: a new positional memory architecture referred to as an Introspection Network, the Neural Turing Machine (NTM) \citep{graves2014neural}, the Least-Recently Used access mechanism (LRU) \citep{santoro2016one}, and the Differentiable Neural Computer (DNC) \citep{graves2016hybrid}.
\item We show that variational inference makes it easy to train scalable models capable of handling high-dimensional input streams leading to new state-of-the-art temporal VAEs.
\item We show that our new models outperform the current state-of-the-art for GTMs based on several tasks that range from generative variants of the copy task to one-shot recall across long time delays.
\item We show that our GTMMs can model realistic 3D environments and demonstrate that these models capture important aspects of physical and temporal consistency, such as coherently generating first-person views under loop closure.  
%
\end{itemize}

\section{Generative Temporal Models}
\label{sect:GTM}
\begin{figure}[t]
    \centering
    \begin{subfigure}[b]{0.24\textwidth}
        \includegraphics[width=\textwidth]{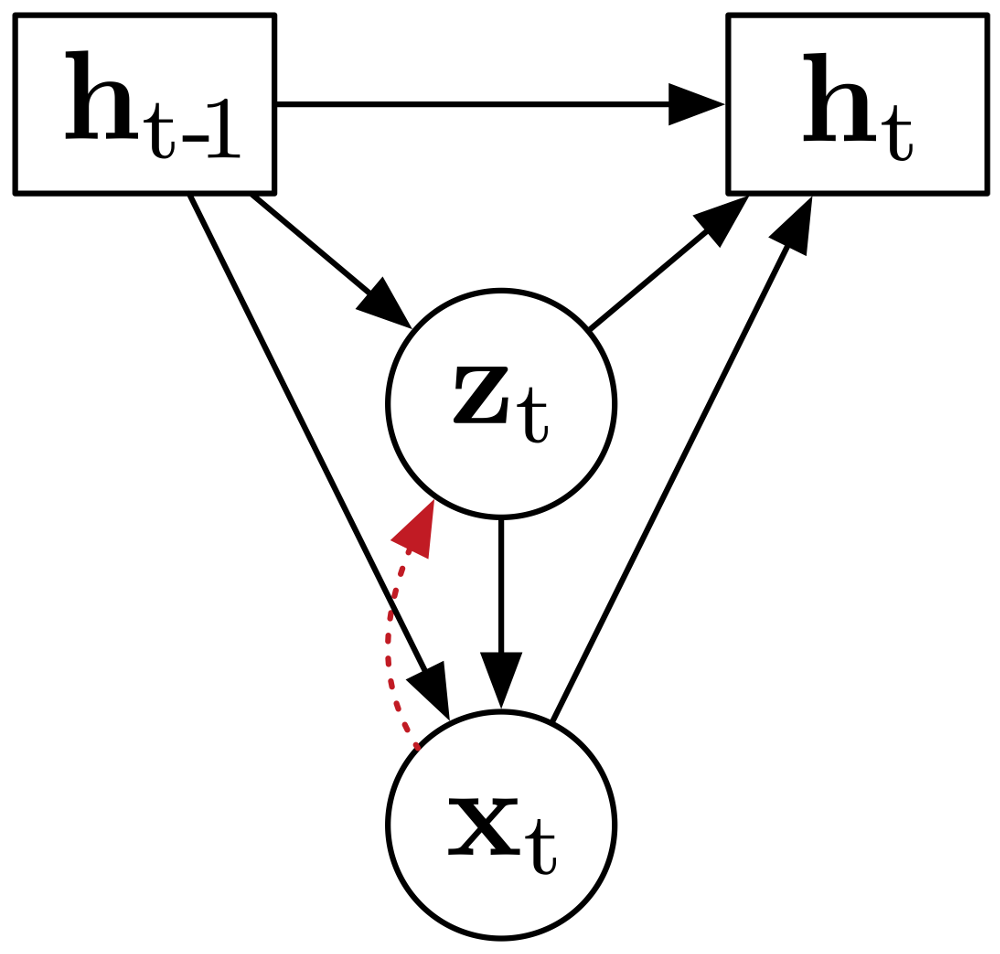}
        \caption{\scriptsize{Variational RNNs}}
        \label{fig:vrnn}
    \end{subfigure}
    \hspace{5mm}
    \begin{subfigure}[b]{0.24\textwidth}
        \includegraphics[width=\textwidth]{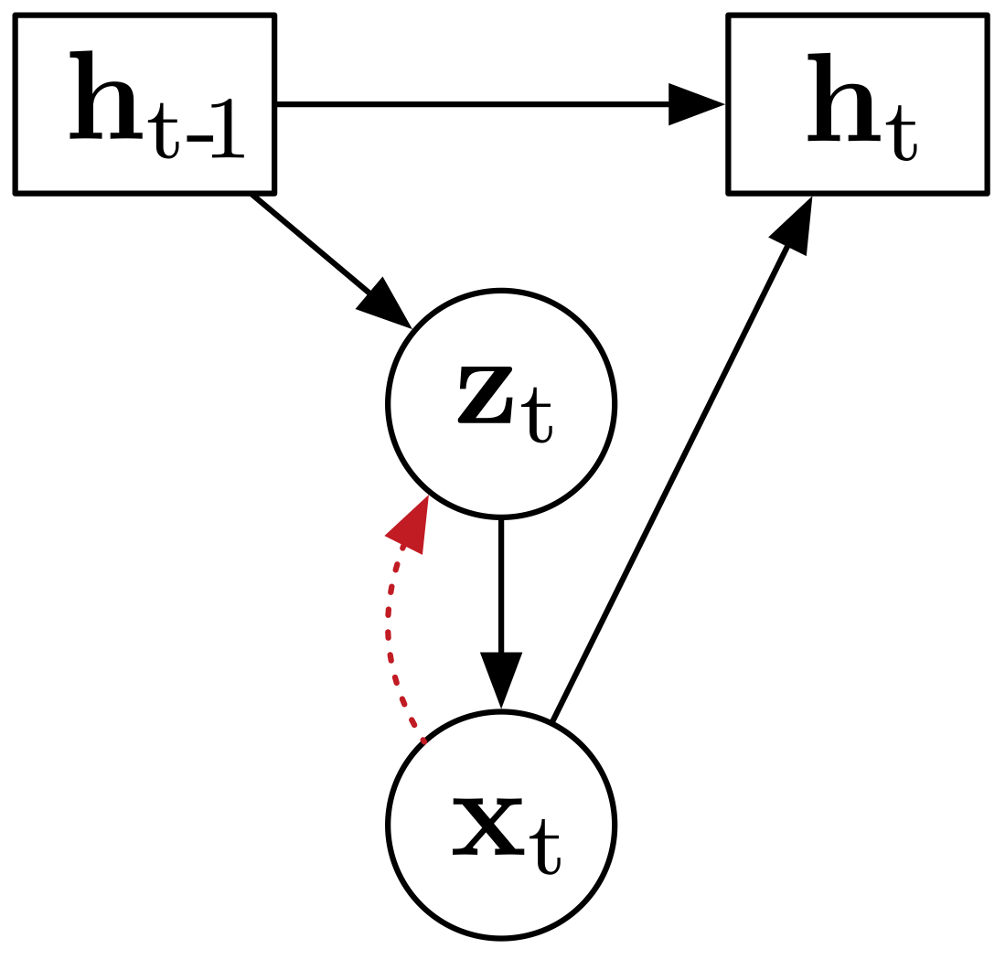}
        \caption{\scriptsize{Autoreg. dynamics}}
        \label{fig:autoreg}
    \end{subfigure}
    \hspace{5mm}
    %
    %
    \begin{subfigure}[b]{0.24\textwidth}
        \includegraphics[width=\textwidth]{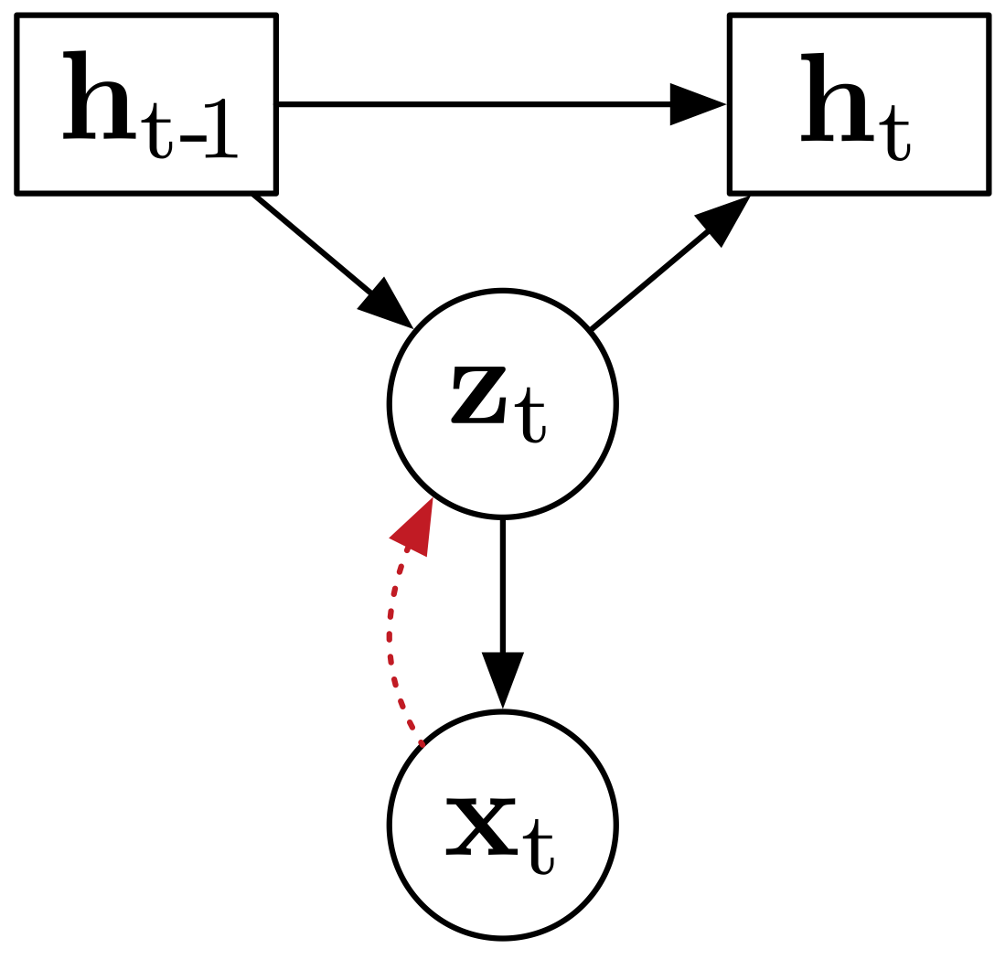}
        \caption{\scriptsize{Latent-only}}
        \label{fig:latent_only}
    \end{subfigure}
    \caption{Variants of generative temporal models. Circled variables are stochastic; boxed variables are deterministic. Solid lines show dependencies in the generative model; dashed lines show additional dependencies for the inference model.}\label{fig:gtm_variants}
\end{figure}

Generative temporal models (GTMs), such as Kalman filters, non-linear dynamical systems, hidden Markov models, switching state-space models, and change-point models \citep{sarkka2013bayesian} are a popular choice for modeling temporal and sequential data using latent variables. These models explain a set of observations $\vx_{\leq T} = \{\vx_1, \vx_2, \ldots, \vx_T\}$ with a set of corresponding latent variables $\vz_{\leq T} = \{\vz_1, \vz_2, \ldots, \vz_T \}$ and specify the joint distribution
\begin{align}
p_\theta(\vx_\seqT, \vz_\seqT) & = \prod_{t=1}^T p_\theta(\vx_t | f_x(\vz_{\leq t}, \vx_\st))p_\theta(\vz_t | f_z(\vz_\st, \vx_\st)),
\label{eq:jointprob}
\end{align}
where $\theta$ are model parameters. This formulation supports a wide range of models, some variants of which are shown in Fig. \ref{fig:gtm_variants}. Particular examples include non-linear state space models \citep{tornio2007time}, Deep Kalman Filters \citep{krishnan2015deep}, and stochastic recurrent neural networks \citep{fraccaro2016sequential, bayer2014learning}. A particular model can be specified in Eq. \eqref{eq:jointprob} by fixing the distributions and the functional dependencies on the conditioned variables. 
\begin{itemize}[leftmargin=*]
\item \textbf{Distributions.} We typically assume that the prior distribution $p_\theta(\vz_t | f_z(\vz_\st, \vx_\st))$ is a Gaussian and the likelihood function $p_\theta(\vx_t | f_x(\vz_{\leq t}, \vx_\st))$ is any distribution appropriate for the observed data, such as a Gaussian for continuous observations or a Bernoulli for binary data.
\item \textbf{Conditional dependencies.}
Our models introduce a deterministic hidden-state variable $h_t$ that is modified at every time point using a \textit{transition map} $\vh_t = f_h(\vh_{t-1}, \vx_t, \vz_t)$.
The function $f_z(\vh_{t-1})$ is a \textit{prior map} that describes the non-linear dependence on past observations and latent variables, using the hidden state, and provides the parameters of the latent variable distribution. 
The non-linear function $f_x(\vz_t, \vh_{t-1})$ is an \textit{observation map} that provides the parameters of the likelihood function, and depends on the latent variables and state. 
These functions are specified using deep neural networks, which can by fully-connected, convolutional, or recurrent networks. 
\end{itemize}

The most general model retains all possible dependencies between latent variables and deterministic state variables in its maps: the transition map $\vh_t = f_h(\vh_{t-1}, \vx_t , \vz_t)$ is parameterised by an LSTM network and depends on the history variable $\vh_{t-1}$, the current observation $\vx_t$, and the current latent variable $\vz_t$; the observation map $f_x(\vh_{t-1}, \vz_t)$ depends on the past history and the current latent variable. This is the structure used by \citet{chung2015recurrent} in variational RNNs (VRNN) in figure \ref{fig:vrnn}. VRNNs forms the baseline in our comparisons since it retains all possible dependencies within the model and provides one of the best existing models. Other dependency structures can also be considered, although they are not used in this paper: GTMs with autoregressive dynamics (figure \ref{fig:autoreg}) have a transition map that depends only on visible variables, i.e. $\vh_t = f_h(\vh_{t-1}, \vx_t)$, and other state-space models use observation maps that depends only on latent variables $p_\theta(\vx_t |  \vx_{<t}, \vz_{\leq t}) = f_x(\vz_t)$.


\subsection{Variational Inference for GTMs}
\label{subsect:VGTM}
Having specified a model \eqref{eq:jointprob}, our task is to infer the posterior distribution of the latent variables and learn the model parameters. Variational inference is currently one of the most widely-used approaches, since it is well suited to problems with high-dimensional observations and high-dimensional parameter spaces. Variational inference also allows for the design of fast and scalable algorithms, is easily composed with other gradient-based learning systems, and provides tools for principled model evaluation and comparison. To compute the marginal probability of the observed data $p(\vx_\seqT)$, we must integrate out any latent variables $\vz_\seqT$. This integration is often intractable and variational methods compute marginal probabilities by transforming this intractable integration problem into a tractable optimization problem. 
We construct a variational bound on the log-marginal likelihood as follows:
\begin{align}
\log p(\mathbf{x}_\seqT) & = \log \int p_\theta(\vx_\seqT, \vz_\seqT) d\vz_\seqT = \log \mathbb{E}_{q_\phi(\vz_\seqT| \vx_\seqT)}\left[\frac{p_\theta(\vx_\seqT, \vz_\seqT)}{q_\phi(\vz_\seqT| \vx_\seqT)}\right] \label{eq:ISestimate}\\
 & \ge \E_{q_\phi(\vz_\seqT|\vx_\seqT)} \left[\log p_\theta(\mathbf{x}_\seqT| \mathbf{z}_\seqT)\right] - \textrm{KL}\left[q_\phi(\mathbf{z}_\seqT|\mathbf{x}_\seqT) \| p_\theta(\vz_\seqT)\right] = \mathcal{F}(q; \vtheta). \label{eq:generalFE}
\end{align}
In Eq. \eqref{eq:ISestimate}, we rewrote the expectation in terms of a distribution $q_\phi(\mathbf{z}_\seqT|\mathbf{x}_\seqT)$ with variational parameters $\vphi$. In equation \eqref{eq:generalFE}, by application of Jensen's inequality, we obtained a lower bound on the marginal likelihood; $KL[q \| p]$ is the Kullback-Leibler divergence between distributions $q$ and $p$. This lower bound \eqref{eq:generalFE}, known as the negative free energy, has two terms that trade off reconstruction accuracy (the expected log-likelihood term) against the complexity of the posterior approximation (the KL-divergence term), and provides a tractable objective function for optimization. 
In this form, the distribution $q_\phi(\mathbf{z}_\seqT|\mathbf{x}_\seqT)$ is an approximation to the true posterior distribution over the latent variables $p_\theta(\mathbf{z}_\seqT|\mathbf{x}_\seqT)$. 

We further choose an auto-regressive form for this distribution. 
\begin{eqnarray}
& q_\phi(\vz_\seqT| \vx_\seqT) = \prod_{t=1}^T q_\phi(\mathbf{z}_{t}|\mathbf{z}_\st, \mathbf{x}_{\leq t}); \quad 
q_\phi(\vz_{\st} | \vx_\st) = \prod_{\tau=1}^{t-1} q_\phi(\vz_{\tau}|\vz_{<\tau}, \vx_{\le \tau}). 
\end{eqnarray}
This choice of approximate posterior distribution allows us to rewrite the total free energy $\mathcal{F}$ as the sum of per-step free energies $\mathcal{F}_t$:
\begin{eqnarray}
& \mathcal{F}(q; \vtheta) = \sum^T_{t=1} \E_{q_\phi(\vz_{\st} | \vx_\st)}\left[ \mathcal{F}_t \right(q; \vtheta)] \label{eq:FEtot} \\
& \mathcal{F}_t = \E_{q_\phi(\vz_{t}|\vz_\st, \vx_\seqt)} \left[ \log p_\theta(\vx_t| \vz_\seqt, \vx_\st) \right] - \text{KL} \left[ q_\phi(\vz_{t}|\vz_\st, \vx_\seqt) \| p_\theta(\vz_t| \vz_\st, \vx_\st) \right] \label{eq:FE}
\end{eqnarray}
A detailed derivation appears in Appendix \ref{appdx:seqVIeqns}. 

%

Recent approaches for variational inference use two additional tools to optimize the free energy. First, since the expectations in \eqref{eq:FEtot} and \eqref{eq:FE} are typically not known in closed form, the gradient of \eqref{eq:FE} is computed using a Monte Carlo estimator. For continuous latent variables, the pathwise derivative (reparameterisation trick) can be used \citep{fu2005stochastic, rezende2014stochastic, kingma2014stochastic}. Second, the approximate posterior distribution $q$ is represented by an inference (or recognition) model whose outputs are the parameters of the posterior distribution. Inference networks amortise the cost of inference across all posterior computations and make joint optimisation of the model and the variational parameters possible. The inference model $q_\phi(\vz_t | f_q(\vx_{\leq t}, \vz_{< t}))$ uses a \textit{posterior map} $f_q$ specified by a deep network that provides the parameters of the $q$-distribution as a function of the current observation, and the past history of latent variables and observations. Latent variable models trained using amortised variational inference and Monte Carlo gradient estimation are referred to as variational auto-encoders (VAEs) \citep{kingma2014stochastic}. For generative temporal models, they are referred to as temporal VAEs.

\section{Generative Temporal Models with External Memory}
\label{sect:vrnn_ntm}
In existing models, temporal structure is captured by LSTM networks with state variables $\vh_t$ (Fig. \ref{fig:gtm_variants}). As we summarised in the introduction and will exhibit in the experiments, LSTMs are powerful sequence models but suffer from the limitation that they strongly couple memory capacity with recurrent processing and the number of trainable parameters. This limitation can result in slow learning or demand large models to achieve high capacity memory. 
To overcome this issue, we now develop generative temporal models with memory (GTMMs), i.e.\ ones that are augmented with external memory systems.

We modify our temporal VAEs to rely on the output of an external memory system, which at every point in time is queried to produce a memory context $\vPsi_t$. The prior and the posterior used become:
\begin{eqnarray}
\textrm{Prior} & p_\theta(\vz_t| \vz_\st, \vx_\st) = \mathcal{N}(\vz_t |f_z^\mu(\vPsi_{t-1}), f_z^\sigma(\vPsi_{t-1})) \\
\textrm{Posterior} & q_{\phi}(\vz_t|\vz_\st,\vx_\seqt) = \mathcal{N}(\vz_t |f_q^\mu(\vPsi_{t-1}, \vx_t), f_q^\sigma(\vPsi_{t-1}, \vx_t))
\label{eq:introspPrior}
\end{eqnarray}
where we use a prior that is a diagonal Gaussian that depends on the memory context through the prior map $f_z$, and use a diagonal Gaussian approximate posterior that depends on the observation $\vx_t$ and the memory context $\vPsi_{t-1}$ through a posterior map $f_q$. We show a stochastic computational graph for the modified generative process in Fig. \ref{fig:stochcompgraph}. This structure is generic and flexible and allows any type of memory system to be used, allowing the remainder of the system to be unchanged since all dependencies are through the memory context $\vPsi_t$. 

External memory systems comprise two components: an \textit{external memory} $\vM_t$, which stores latent variables $\vz_t$ (or transformations of them), and a \textit{controller}, which implements the addressing scheme that informs memory storage and retrieval. Two types of addressing schemes are possible: \textit{content-based addressing} accesses memories based on their similarity to a given cue, while \textit{position-based addressing} accesses memories based on their position within the memory-store. We now expand on four types of memory systems that have different characteristics, describing the specific memory and controllers used, and how the final memory context $\vPsi_t$ is computed.

\subsection{Introspective-GTMMs}
\label{sect:introspection}
\begin{figure}[t]
	\vspace{-0.1in}
	\centering
	\includegraphics[height=0.22\linewidth]{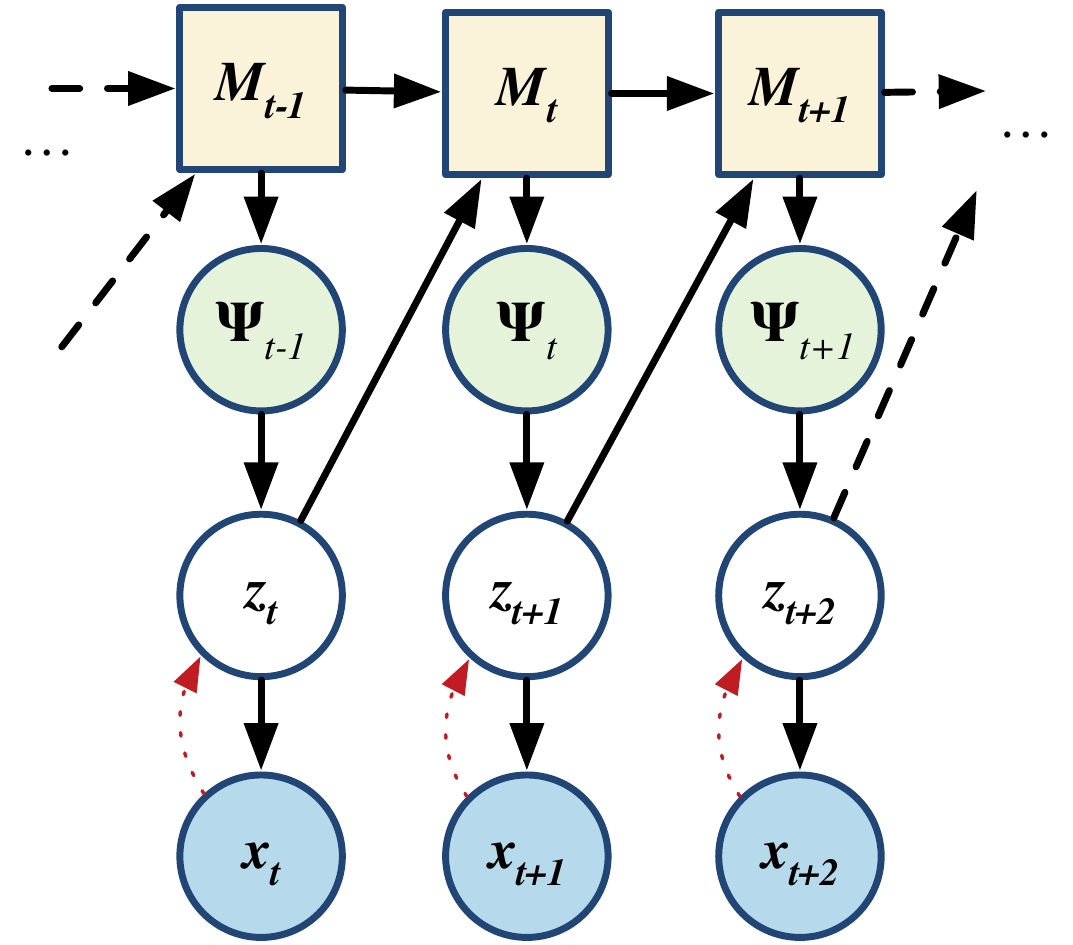}
	\hspace{1mm}
	\includegraphics[height=0.22\linewidth]{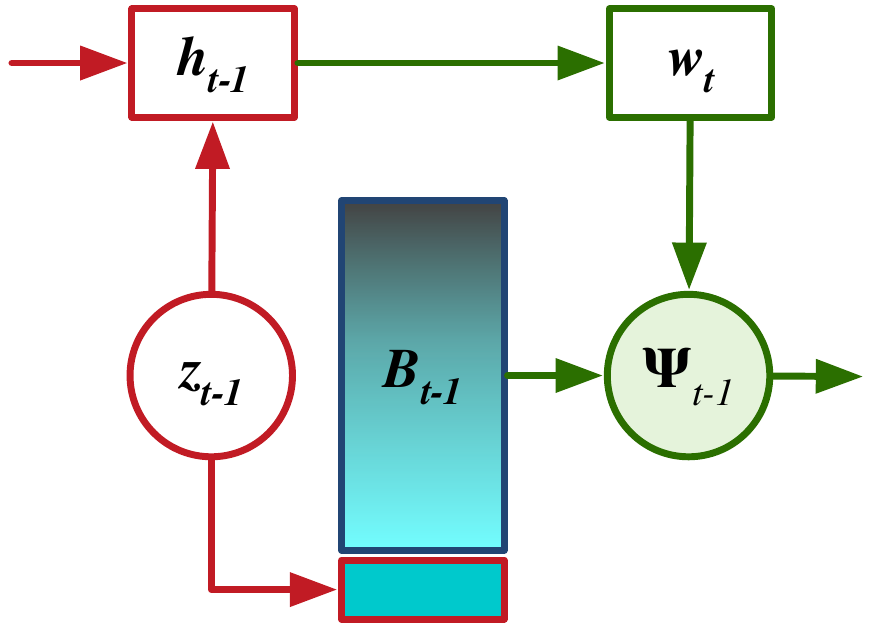}
	\hspace{1mm}
	\includegraphics[height=0.22\linewidth]{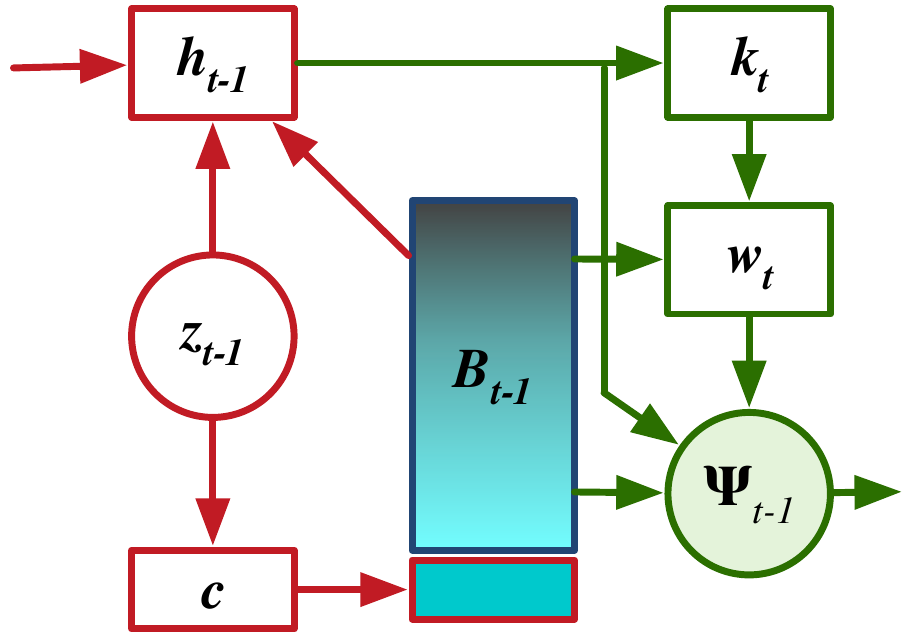}
	\caption{Components of a generative temporal model with external memory. (Left) High-level structure of the model showing the memory system M and how it connects to the generative model. Red and green lines indicate writing and reading operations, respectively. At update $t$, the controller state from time $t-1$ is combined with the latent variable from time $t-1$ to produce the attention weight. This produces a memory context that is only a function of the data that were in memory before time $t$, so we denote it $\vPsi_{t-1}$. (Middle) Schematic of the introspective memory system. (Right) Schematic of memory systems like NTM and DNC.}
    \label{fig:stochcompgraph}
    \vspace{-0.2in}
\end{figure}
We now develop an external memory system that uses a position-based addressing scheme for fast learning of temporal dependencies, related to the Pointer Networks of  \cite{vinyals2015pointer}. It is able to effectively handle sequences with temporally-extended dependency structures, trains quickly, and can be robustly applied to a wide variety of tasks (and we verify this in the experimental section). We refer to this memory system as an introspection network.

\textbf{Memory.} The memory $\vM$ is a first-in-first-out buffer with at most $L$ storage locations into which latent variables $\vz_t$ are written as they are generated at each time step. It is natural to directly store the latent variables, since they are compressed representations of the data at each time point. 
This type of memory does not require the model to learn how to \textit{write} to memory, but only how to read from it. This feature is what enables fast learning.

\textbf{Controller.} The controller is responsible for memory retrieval. At every time step $t$, the controller first updates the hidden states using an LSTM network $f_{rnn}$ (Eq. \ref{eq:introsp_stateupd}), which fuses information from the previous hidden state $\vh_{t-1}$ and the previously generated latent variable $\vz_{t-1}$. If additional context information $\vc_{t}$ is available, then this is also included as an input. To access a memory, soft-attention weights are computed using an attention network $f_{att}$ based on the output of the controller $\vh_t$ (Eq. \ref{eq:introsp_attent}). A set of $R$ attention weights (or read heads) is used to retrieve multiple memories at the same time.
Each attention weight $\vw^{r}_t$ is used to compute a weighted average over the rows of the memory matrix to produce a retrieved memory vector $\vphi^r_t$. 

\begin{eqnarray}
\textrm{State update} & \vh_t = f_{rnn}(\vh_{t-1}, \vz_{t-1}, \vc_{t}) & \label{eq:introsp_stateupd}\\
\textrm{Attention} & \vw^{r}_t = f_{att}(\vh_{t}); \quad \|\vw^{r}_t\|=1, w^r_{t}[i] >0 & \label{eq:introsp_attent}\\
\textrm{Retrieved memory} & \vphi^r_t = \vw^r_t \cdot \vM_{t-1} & \label{eq:introsp_retrmem}
\end{eqnarray}

A number of attention functions $f_{att}$ can be used, including softmax and Gaussian. We make use of normalised linear attention with softplus outputs $\mathbf{w}_{t} = {k(\mathbf{h}_{t})}/\sum_u {k(\mathbf{h}_{u})}$, where the function $k$ is a deep feed-forward network. This attention system proved easy to use and did not require special initialisation. In this model, we found that softmax attention had slower convergence. 

\textbf{Gating mechanism.} The ability to retrieve multiple memories $\vphi_t^r$ makes it possible for the latent variables $\vz_{t}$ to depend on a variable number of past latent variables. We allow the network to adjust the importance of each of the retrieved memories $\vphi_t^r$ by learning correction biases $\vg_t^r$. The corrections are passed through a sigmoid function $\sigma(\cdot)$ and element-wise multiplied with the context vector.

\begin{eqnarray}
\textrm{Memory context} & \vPsi_t^r = \vphi_t^r \odot \sigma(\vg^r_{t-1}) &  \label{eq:introsp_memcontext}
\end{eqnarray}

The final memory context $\vPsi_t$ that is the output of the memory system is the concatenation of the memory-contexts for each read-head, $\vPsi_t = [\vPsi_t^1, \vPsi_t^2, \ldots, \vPsi_t^R]$, and forms the memory context that is passed to the generative model. The complete flow of information in the Introspection Network is shown in the stochastic computational graph in Fig. \ref{fig:stochcompgraph}.

\subsection{Models with Content-Based Addressing}
Introspective GTMMs can learn fast, but their simple memory structure limits the range of applications to which they can be applied. We now develop GTMMs with three alternative types of memory architectures: the Neural Turing Machine (NTM) \citep{graves2014neural}, which combines both content-based and positional addressing; Least-Recently Used (LRU) access, which exclusively employs content-based addressing; and the Differentiable Neural Computer (DNC) \citep{graves2016hybrid}, which uses content-based addressing and a mechanism of positional addressing that links positions in memory based on temporal adjacency of writing. We call these models the NTM-GTMM, LRU-GTMM, and DNC-GTMM, respectively. We describe high-level aspects of the memory and controllers used, but defer detailed discussion of the properties and alternative parameterisations to \citet{graves2014neural}, \citet{santoro2016one} and \citet{graves2016hybrid}.

\textbf{Memory.} Unlike the first-in-first-out buffer used previously, the memory for NTMs and DNCs are a generic storage that allows information to be written to, and read from any location. 

\textbf{Controller.}
The controller uses an LSTM network $f_{rnn}$ (Eq. \ref{eq:dnc_stateupd}) that updates the state-history $\vh_t$ and the external memory $\vM_t$ using the latent variables from the previous time step and any additional, context information $\vc_{t}$ on which the generative model is conditioned:

\begin{eqnarray}
\textrm{State update} & (\vh_t, \vM_t) = f_{rnn}(\vh_{t-1}, \vM_{t-1}, \vz_{t-1}, \vc_{t}) & \label{eq:dnc_stateupd}
\end{eqnarray}

To perform a content-based read of $R$ items from the memory $\vM_t$, the controller generates a set of keys $\mathbf{k}_t^r$ \eqref{eq:dnc_keys}, and compares them to each row of the memory $\vM_{t-1}$ using a cosine similarity measure to yield a set of soft attention weights \eqref{eq:dn_weights}. The retrieved memory $\vphi_t^r$ is then obtained by a weighted sum of the attention weights and the memory $\vM_{t-1}$ \eqref{eq:dnc_retmem}.

\begin{eqnarray}
\textrm{Keys} & \mathbf{k}_t^r = f_{key}^r(\vh_{t}); \quad r \in \{1, \dots, R\} & \label{eq:dnc_keys}\\
\textrm{Attention} & \vw^{r}_t = f_{att}(\vM_{t-1}, \mathbf{k}_t); \quad \vw^{r}_t[i] \geq 0 ; \|\vw^{r}_t\|=1 & \label{eq:dn_weights}\\
\textrm{Retrieved memory} & \vphi^r_t = \vw^r_t \cdot \vM_{t-1} & \label{eq:dnc_retmem}
\end{eqnarray}

The memory context $\vPsi_t$ that is passed to the generative model is the concatenation of the retrieved memories for each read-head and the controller state, $\vPsi_t = [\vphi_t^1, \ldots, \vphi_t^R, \vh_t]$.
\section{Evaluation Methodology}
\label{sect:exp}
We evaluated our models both qualitatively and quantitatively. Qualitative assessment involved the visual inspection of generated sequences; for example, if the task were to copy a particular portion of an observed sequence after some set number of steps, then this copy procedure should be evident in sequences generated by the models. Qualitative assessments revealed significant differences between models even when differences in their variational lower bounds were minimal. We also used three quantitative metrics to gather a more complete picture of the model behaviour: first, the variational lower bound objective function, tracked across training; second, the KL-divergence at a particular time-point for every training sequence (the last time-point per episode) tracked across training steps; and third, the per time-point KL-divergences averaged over a batch of sequences after training was completed.

Per time-step KL-divergences, $\text{KL} \left[ q_\phi(\vz_{t}|\vz_\st, \vx_\seqt) \| p_\theta(\vz_t| \vz_\st, \vx_\st) \right]$, measure the number of bits of additional information needed to represent the posterior distribution relative to the prior distribution over the latent variable being used to explain the current observation. 
They indicate the amount of prior knowledge the model contains. 
If a KL-divergence is close to zero, then the current observation is fully-predictable from previous information.
For the tasks considered here, which were defined by random sequences at every episode, this would imply that the model stores information in memory from the beginning of the episode to construct predictive priors for the rest of the episode.

Calculating this quantity across training sequences for the last time point in each sequence (the second metric) demonstrates how quickly the memory system becomes useful for prediction across training (the last time point in our problems was always the most predictable). 
Viewing the average per time-step KL, averaged over a batch of sequences after training (the third metric), indicates how much information a trained model gathers throughout a sequence to make predictions.

\subsection{Training Details}
\label{sect:training}
Our posterior and observation maps used convolutional and deconvolutional networks, in some cases with residual, skip connections \citep{he2015deep}. Refer to Appendix \ref{appen:visualArchitectures} for explicit details. 
All models were trained by stochastic gradient descent on the variational lower bound (Eq. \ref{eq:FE}) using the Adam optimizer \citep{CorrKingma2014} with a learning rate of $10^{-3}$ (except for sequences with $>100$ steps, where we used $10^{-4}$). 
Mini-batches of $10$ training sequences were used for computing gradients in all tasks. 
For tasks involving digits and characters, we used latent variables of size $32$; for the 3D environment, we used latent variables of size $256$. 

We used five read heads in all tasks. The number of memory slots used for the GTMMs was taken to be the same as the number of steps in the training sequences in each task, except for the LRU-GTMM, which used a number of slots that was five times the number of time steps (note: the LRU ties the number of write heads to the number of read heads, and with many write heads it fills up a small memory quickly). The hidden state size of the LSTM in all models was chosen to keep the total number of parameters within $\sim 5\%$ of one another (see Table \ref{table:modelTaskParameters}). 

\begin{table}[t]
\centering
\caption{Number of parameters used in models.}
\begin{tabular}{lcc}
\hline
Model & Digits and characters  & 3D environments  \\
\hline \hline
VRNN & 1,884,177 &  5,912,706  \\
Introspective-GTMM & 1,863,107 & 5,972,806 \\
NTM-GTMM & 1,869,381 & 5,986,692 \\
LRU-GTMM & 1,866,282 & 5,979,115  \\
DNC-GTMM & 1,859,336 & 5,980,327 \\
\hline
\end{tabular}
\label{table:modelTaskParameters}
\end{table}

For each model, we ran 20 replicas with the same hyperparameters. When we performed quantifications for figures, we first averaged over all 10 example sequences in a mini-batch, then we computed means and standard errors across the replicas for the model type.

\section{Experimental Results}
We tested our models on seven tasks that probed their capacity to learn and make predictions about temporal data with complex dependencies. Tasks involved image-sequence modelling, and offered tests of deduction, spatial reasoning, and one-shot generalisation. Example training sequences are provided for each task described below. In Appendix \ref{appen:gen_samples}, we present generated samples for most of the tasks. For tasks with artificial data sets, pseudo-code to generate the training sequences is given in Appendix \ref{pseudocodeForTasks}.

\subsection{Perfect Recall}
\begin{figure}[t]
	\centering
    \includegraphics[width=\linewidth]{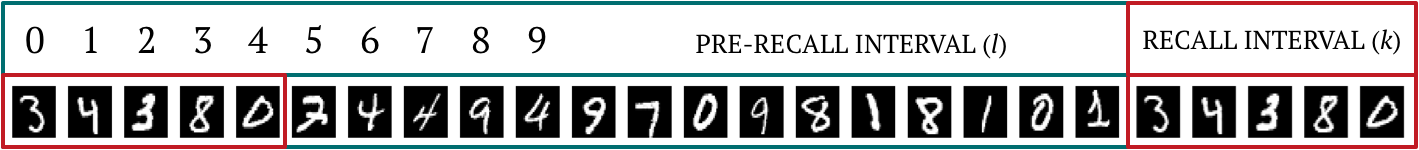}
    \caption{Example sequence for the \emph{perfect recall} task using a pre-recall interval $l=20$ followed by a recall interval $k=5$.}
    \label{fig:perfectRecallExample}
\end{figure}
\begin{figure}[t]
\centering
\includegraphics[width=1\textwidth]{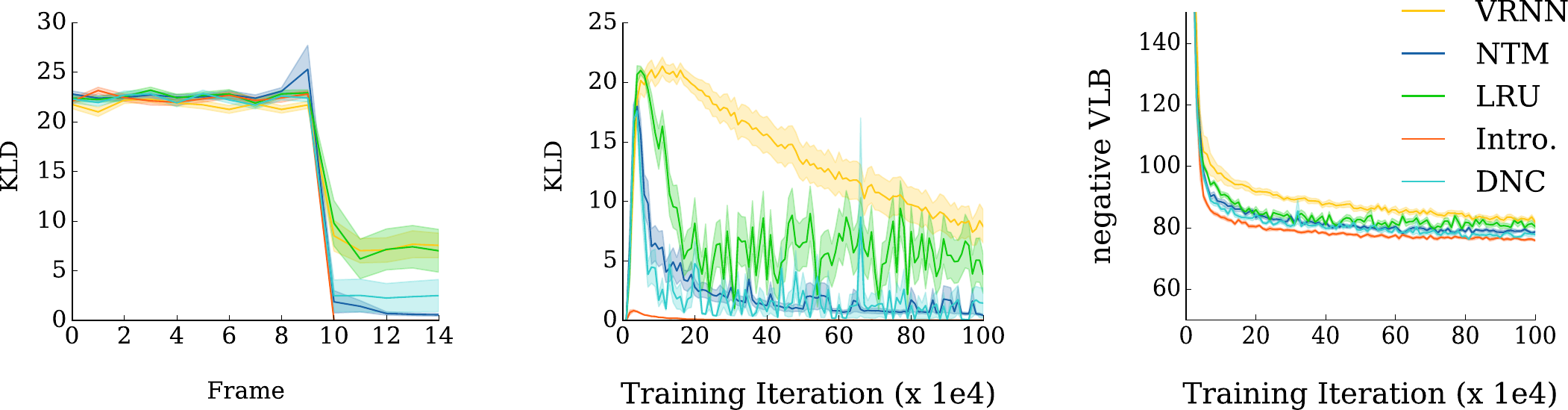}
\includegraphics[width=1\textwidth]{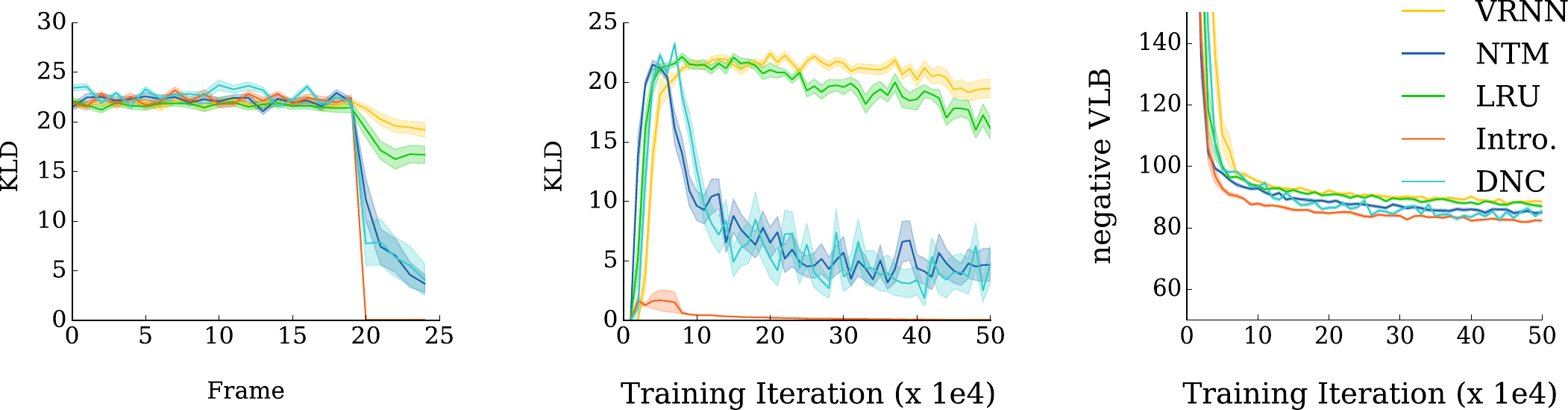}
\includegraphics[width=1\textwidth]{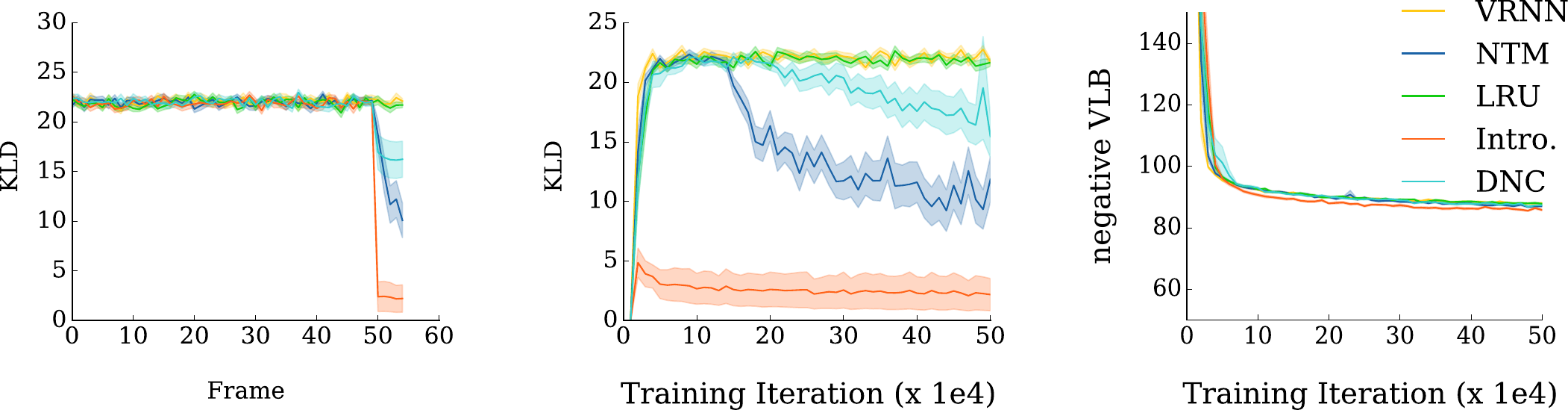}
\caption{\emph{Top Row}: the perfect recall task with $l=15$ and $k=5$. \emph{Left:} The average Kullback-Leibler divergence (KLD) per frame, serving as a measure of prediction error between the prior and posterior. Each of the models learned that there is repetition at frame 10, but the Introspective GTMM exhibited the lowest error. \emph{Middle:} The last frame KLD was also lowest for the Introspective GTMM at convergence. \emph{Right:} The Introspective GTMM convergenced fastest and to the lowest level, but we see that the negative variational lower bound was close for all models.
\emph{Middle Row}: the perfect recall task with $l=20$ and $k=5$. The results were roughly similar.
\emph{Bottom Row}: the perfect recall task with $l=50$ and $k=5$. Over substantially larger time intervals, the models were able to detect regularity in the data sequences.
\label{fig:per_recall_results}}
\end{figure}
Training sequences consisted of $k$ randomly sampled MNIST digits to be remembered during a pre-recall interval, which extended for $l$ time steps. Thus, $l-k$ digits were distractor stimuli. A recall interval occurred after $l$ steps, during which the first $k$ images were presented again. We constructed variants of the task with $k=5$ and $l \in \{15, 20, 50\}$. Fig. \ref{fig:perfectRecallExample} shows a typical training sequence for this task. To succeed at this task, the models had to encode and store the first $k$ images, protect them during the distractor interval, and retrieve them during the recall interval. Successful use of memory would elicit a drop in the KL-divergence during the recall interval, since information stored in memory -- and not information extracted from the current observation -- would be used for image reconstruction.

The performance for this task is reported in Fig. \ref{fig:per_recall_results}. All models showed the effect that the KL-divergence between the prior and posterior becomes reduced at the beginning of the recall phase. However, for the GTMMs and especially the Introspective GTMM, the reduction was most significant. This implies that the models were predicting the arriving frames based on memory. The effect was more pronounced the larger the sequence length.

\subsection{Parity Recall}
\begin{figure}[t]
	\centering
    \includegraphics[width=\linewidth]{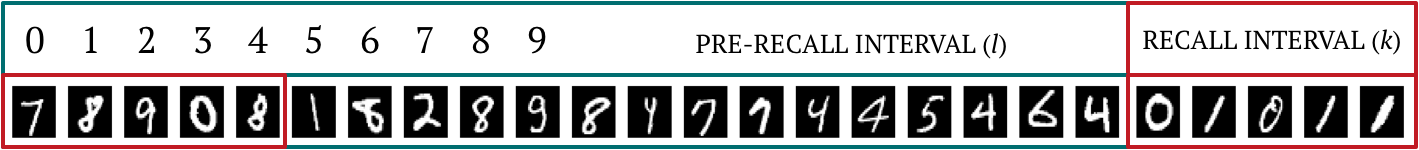}
    \caption{In the \emph{parity recall} task, the recall interval consisted of images indicating the evenness or oddness of each of the initial $k$ images.}
    \label{fig:parityRecallExample}
\end{figure}
\begin{figure}[t]
	\centering
	\includegraphics[width=1\textwidth]{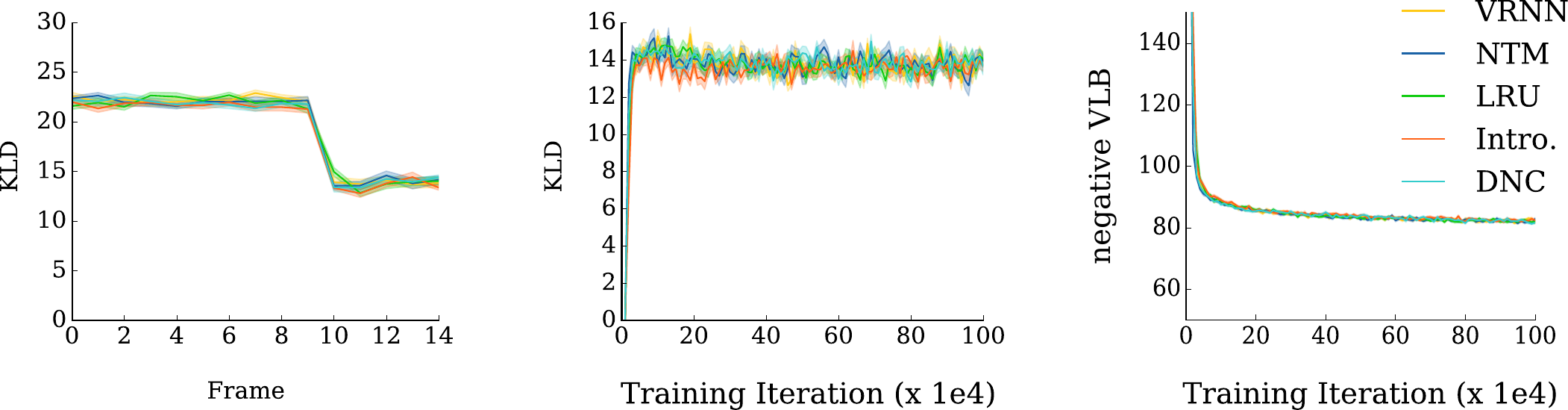}
	\includegraphics[width=1\textwidth]{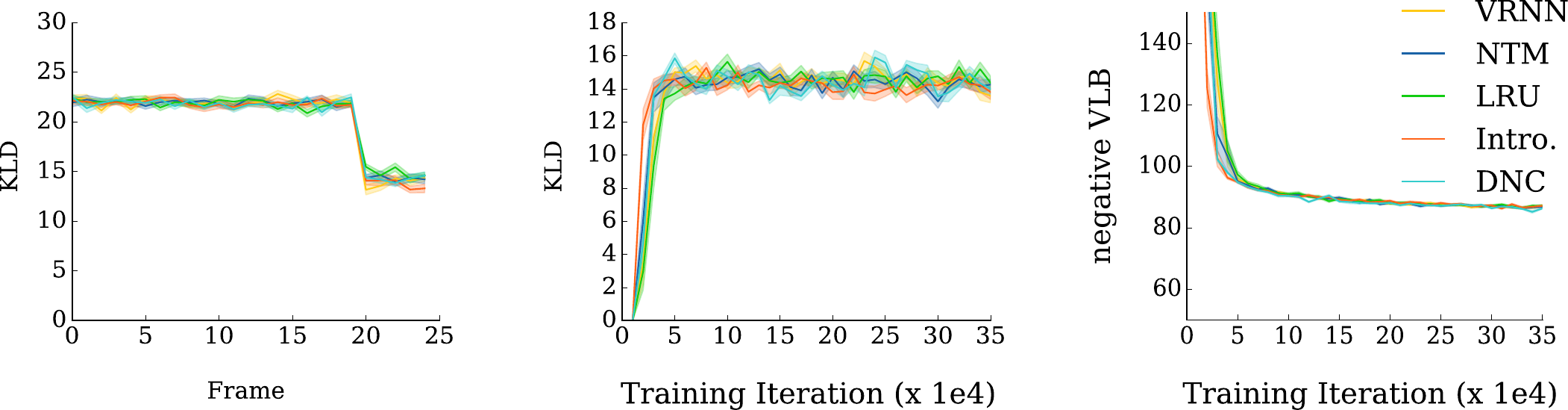}
	\includegraphics[width=1\textwidth]{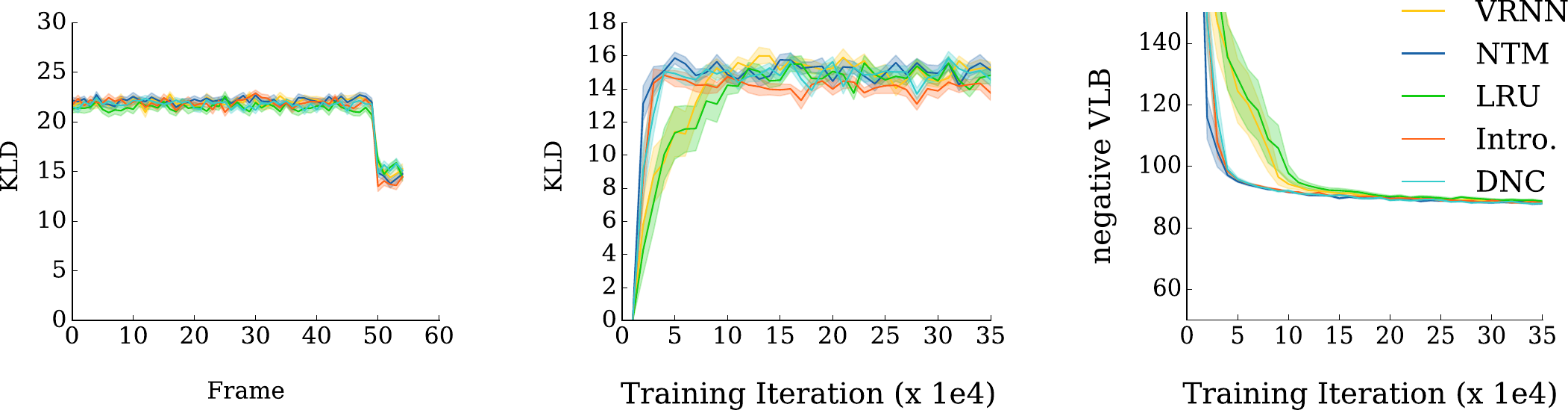}	
    \caption{\emph{Top Row}: the parity recall task with $l=15$ and $k=5$. The difference between models was minimal here, likely because the memory storage required to solve the task is only $k=5$ bits.
    \emph{Middle Row}: the task with parameters $l=20$ and $k=5$.
    \emph{Bottom Row}: with $l=50$ and $k=5$. Again, all models succeeded equally despite the longer delay.    
    \label{fig:par_recall_results}}
\end{figure}
In contrast to the previous experiment in which exact recall of the images was demanded, in the parity recall task we asked the model to identify and report on a latent property of the data. During the recall interval, the model must generate a sequence of $k$ $0$-s and $1$-s, matched to the parity of the first $k$ images. That is, the first recalled digit should be a zero if the first digit in the initial sequence was odd, and one if it was even. Fig. \ref{fig:parityRecallExample} shows a typical training sequence. Successful models, then, need to implicitly classify input digits. Although the \textit{computation} required for this task is more complicated than for perfect recall, the information content that is to be stored is actually smaller -- i.e., a single bit per image. 

Less memory is required for parity recall than for perfect recall as the details of each image need not be retained; instead, only the parity of each image should be tracked, requiring 5 bits total. All models performed satisfactorily here, as we see in Fig. \ref{fig:par_recall_results}, exhibiting KL reductions when the number of possible digit classes drops from 10 to only 1 (the remembered parity digit images). The models were able to contend with long delays equally, suggesting that the primary advantage for GTMMs over VRNNs is in tasks that require the storage of a large number of bits.

\subsection{One-Shot Recall}
\begin{figure}[t]
	\centering
    \includegraphics[width=\linewidth]{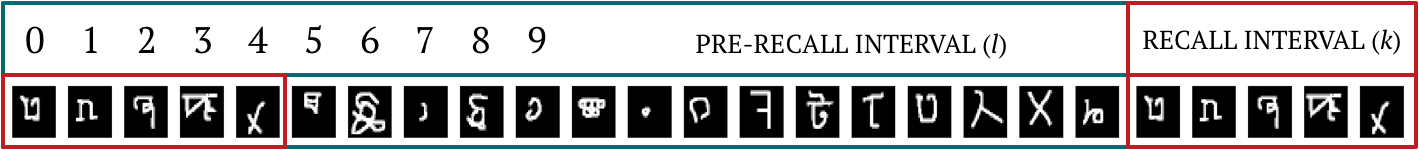}
    \caption{For the \emph{one-shot recall} task, sequences at test time were created from a set of characters that were not used during training. Even so, perfect recall should still be possible.}
    \label{fig:OmniRecallExample}
\end{figure}
\begin{figure}[t]
	\centering
	\includegraphics[width=1\textwidth]{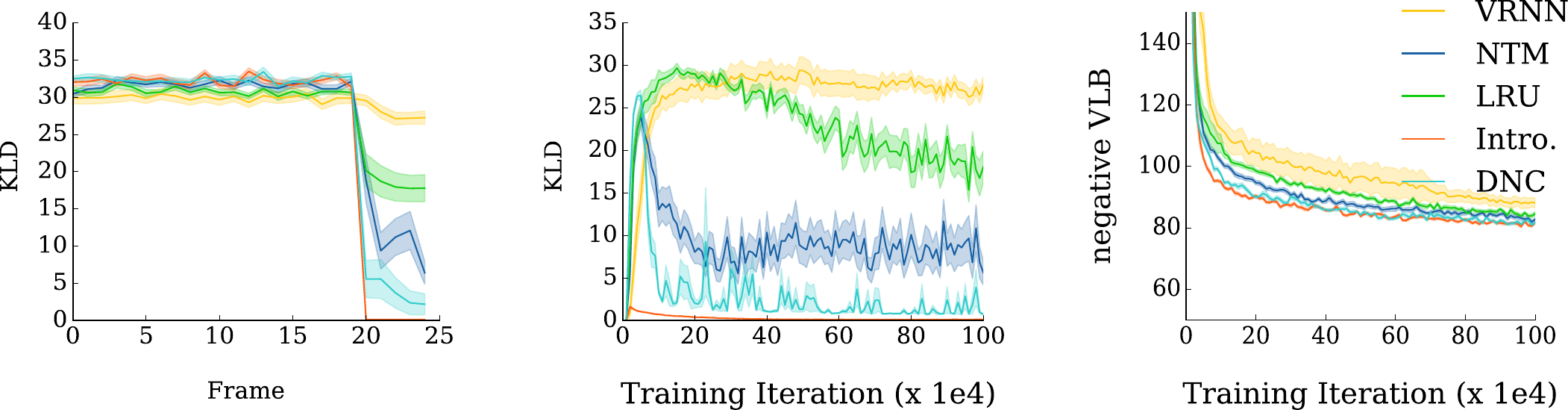}
    \caption{The one-shot recall task with experimental parameters $l=20$ and $k=5$. Again, because the task was very similar to perfect recall in structure, the models performed in a comparable rank order, with the Introspective GTMM showing significant KL reductions when the sequence was predictable from memory.
    \label{fig:omni_recall_results}}
\end{figure}
We also examined the abilities of the GTMMs to memorise novel information by testing on sequences of data on which they had not been directly trained. A typical training sequence was shown in Fig. \ref{fig:OmniRecallExample}, where the images at every point in time are drawn from the Omniglot data set \citep{lake2015human}. The training data consisted of all 50 alphabets with three characters excluded from each alphabet. The unseen characters were used to form new, unseen sequences at test time. The task was otherwise the same as the perfect recall task, but the demands on the generative model and memory were more substantial.  

The GTMMs all outperformed the VRNN with the Introspective GTMM showing significant reductions in KL divergence at the beginning of the recall phase (Fig.  \ref{fig:omni_recall_results}). This was notable because the memorised images are entirely novel hold-outs from the training set.
Thus, the GTMMs, and in particular the Introspective GTMM, were able to construct useful latent variable representations for novel stimuli, store them in memory, and use them to predict future events during the recall phase.

\subsection{Learning Dynamic Dependencies}
\begin{figure}[t]
	\centering
    \includegraphics[width=\linewidth]{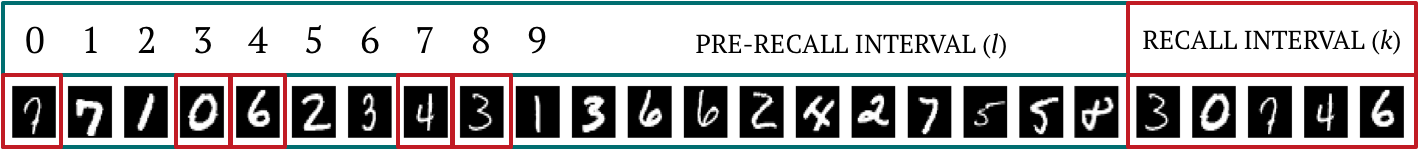}
    \caption{Training sequence for the \emph{dynamic dependency} task following an index-and-recall game in which each image digit provided a positional reference to the next digit in the sequence order.}
    \label{fig:dynDepExample}
\end{figure}
\begin{figure}[t]
	\centering
	\includegraphics[width=1\textwidth]{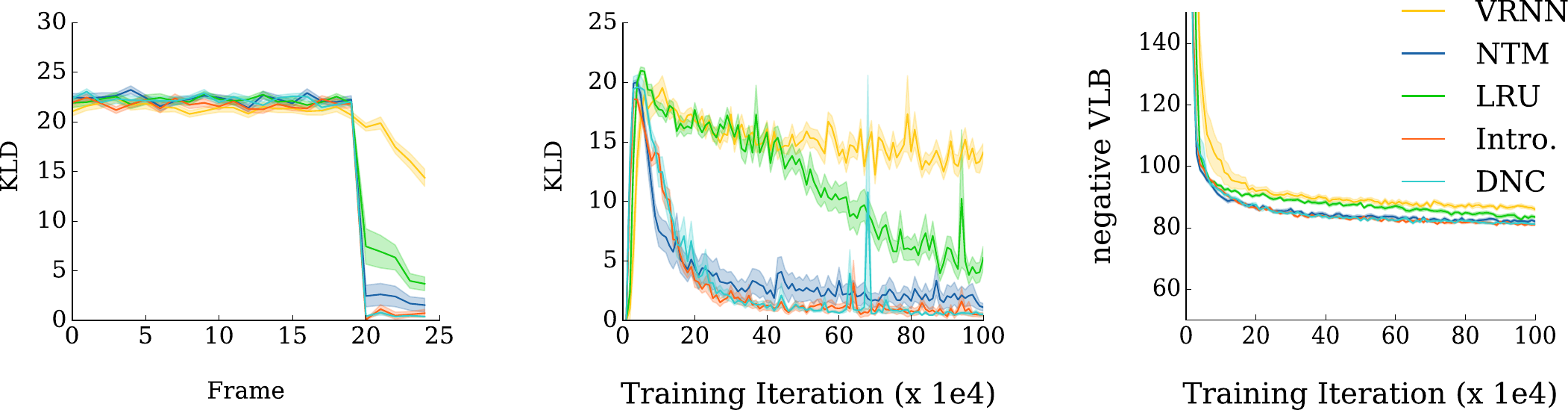}
    \caption{Dynamic dependency task with $l=20$ and $k=5$: The DNC-GTMM and the Introspective GTMM were the best at this complex addressing task.
    \label{fig:dyn_dep_results}}
\end{figure}
The preceding tasks have all demanded ordered recall of the sequence.
Here, we tested whether recall in a more complicated order is possible.
A typical training sequence is shown in Figure \ref{fig:dynDepExample}. We began with a sequence of $l$ digits as before. 
The next $k$ digits were generated by an ``index-and-recall'' game. 
In the figure example, the final digit in the pre-recall interval is an 8.
The numerical value of the digit indicates from which position in the sequence the next digit is copied. Here, position 8 contains a 3, which is the first digit in the recall interval. Position 3 contains a 0, and so on. Successful models therefore had to learn to classify digits and to use the class labels to find images based on their temporal order of presentation.

This task requires the models to learn an algorithmic addressing procedure in which the current image indicates the time point the next image was stored, allowing the memory address storing the latent variables from that time point to be looked up. All of the GTMMs perform considerably better than the VRNN on the task, with the most substantial improvements achieved by the Introspective GTMM and the DNC-GTMM (Fig. \ref{fig:dyn_dep_results}).

\subsection{Similarity-Cued Recall}
\begin{figure}[t]
	\centering
    \includegraphics[width=\linewidth]{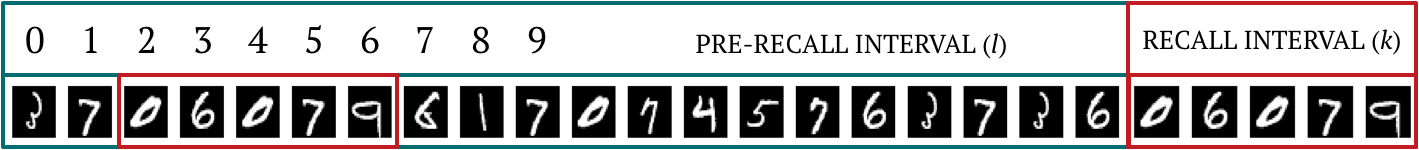}
    \caption{Example training sequence for the \emph{similarity-cued recall} Task.}
    \label{fig:contentDepExample}
\end{figure}
\begin{figure}[t]
	\centering
	\includegraphics[width=1\textwidth]{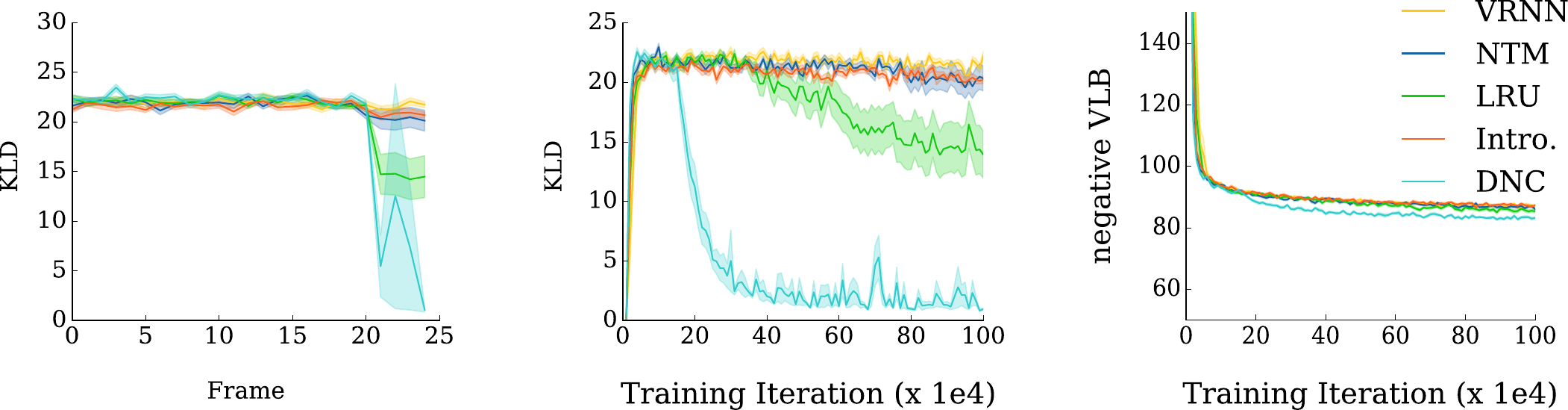}
    \caption{Similarity-cued recall with temporal dependencies of length $l=20$ and $k=5$. The memory models with content-based addressing, i.e., the NTM-, LRU-, and DNC-GTMM, showed the most significant KL reductions and lowest task losses.
    \label{fig:content_results}}
\end{figure}
The last task probed positional indexing, but here we construct a task that demands content-based addressing.
In each training sequence, we first present a \emph{random sequence} of digits for $l$ time steps. The $k$ digits in the \emph{recall interval} are a randomly chosen, contiguous sub-sequence of length $k$ from the pre-recall interval (Fig. \ref{fig:contentDepExample}).
To solve this task, a model must be able to use the first image in the recall interval as a cue, find the most similar image to the cue that it has seen previously, and produce the temporal sequence that followed it.

Similarity-cued recall played more strongly to the advantages of memory systems with content-based addressing. The task required using a cue image to find the images in sequence that followed the cue during the pre-recall exposure phase. To perform this operation, the memory systems with content-based addressing, the NTM, LRU, and DNC, could encode the cue and look up similar feature encodings in memory. As long as there was a mechanism to iterate through the subsequent latent variables, the task can then be easily solved. DNC-GTMM could use its temporal transition links for this task, as described in \citep{graves2016hybrid}. Because it lacks content-based addressing, the Introspective-GTMM did not perform better than the VRNN (Fig. \ref{fig:content_results}).

\subsection{Navigation in an MNIST Map}
\begin{figure}[t]
	\centering
    \includegraphics[width=\textwidth]{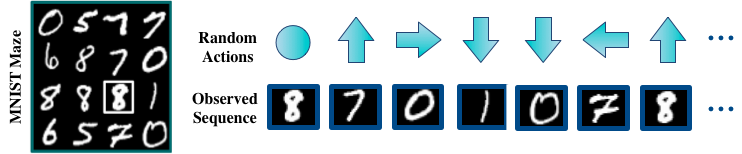} 
    \caption{A random walk action sequence was provided to the model along with images found at each location. At boundaries, the action sequence was restricted to stay in bounds. Each action was provided as a conditioning variable for generation that was not modeled itself.}
    \label{fig:mnist_maze}
\end{figure}
An important motivation for developing GTMMs was the desire to improve the capacity of agents to understand the spatial structure of their environments. 
As an example problem, we created a 2D environment represented by a $4 \times 4$ grid, where each grid cell contained a random MNIST digit (Fig. \ref{fig:mnist_maze}). 
In this case, instead of an agent, we took a random walk on the grid using actions up, down, left, and right for 25 steps to move between neighbouring locations, receiving an observation corresponding to the current grid cell. The actions were always treated as context variables and were not predicted by the GTMMs.
We expected that the GTMMs conditioned on random walk action sequences would be able to generate the same observation in case a grid cell is revisited in an action sequence; that is, the GTMM should have maintained a coherent map of an environment.

\begin{figure}[t]
	\centering
	\includegraphics[width=1.\textwidth]{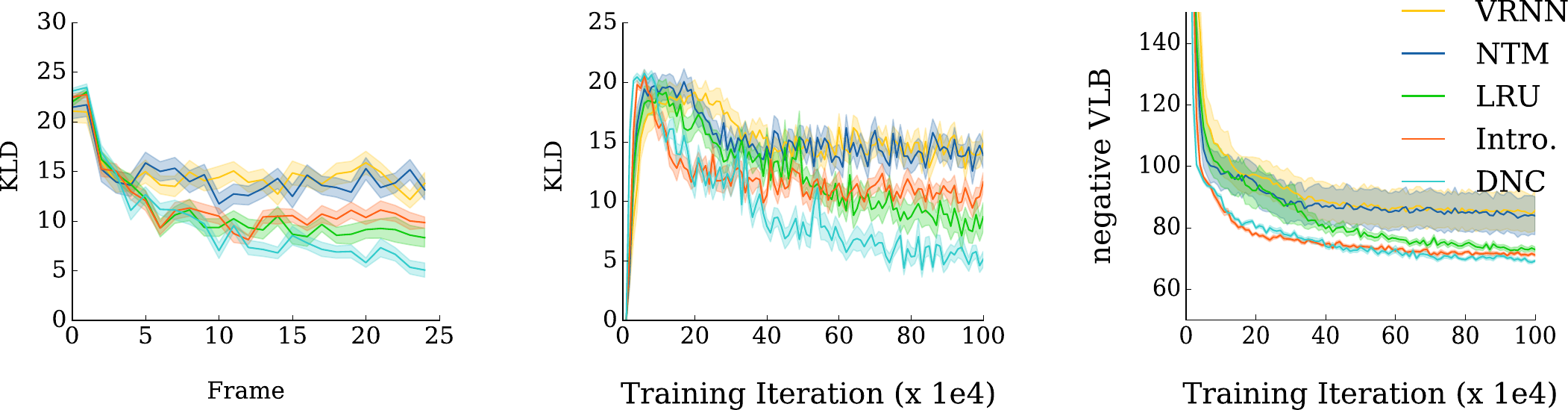}
    \caption{MNIST Maze task. The GTMMs with content-based addressing acquired information about the environment and could coherently model it.
    \label{fig:mnist_maze_results}}
\end{figure}
\begin{figure}[p]
	\centering
	\includegraphics[width=1.\textwidth]{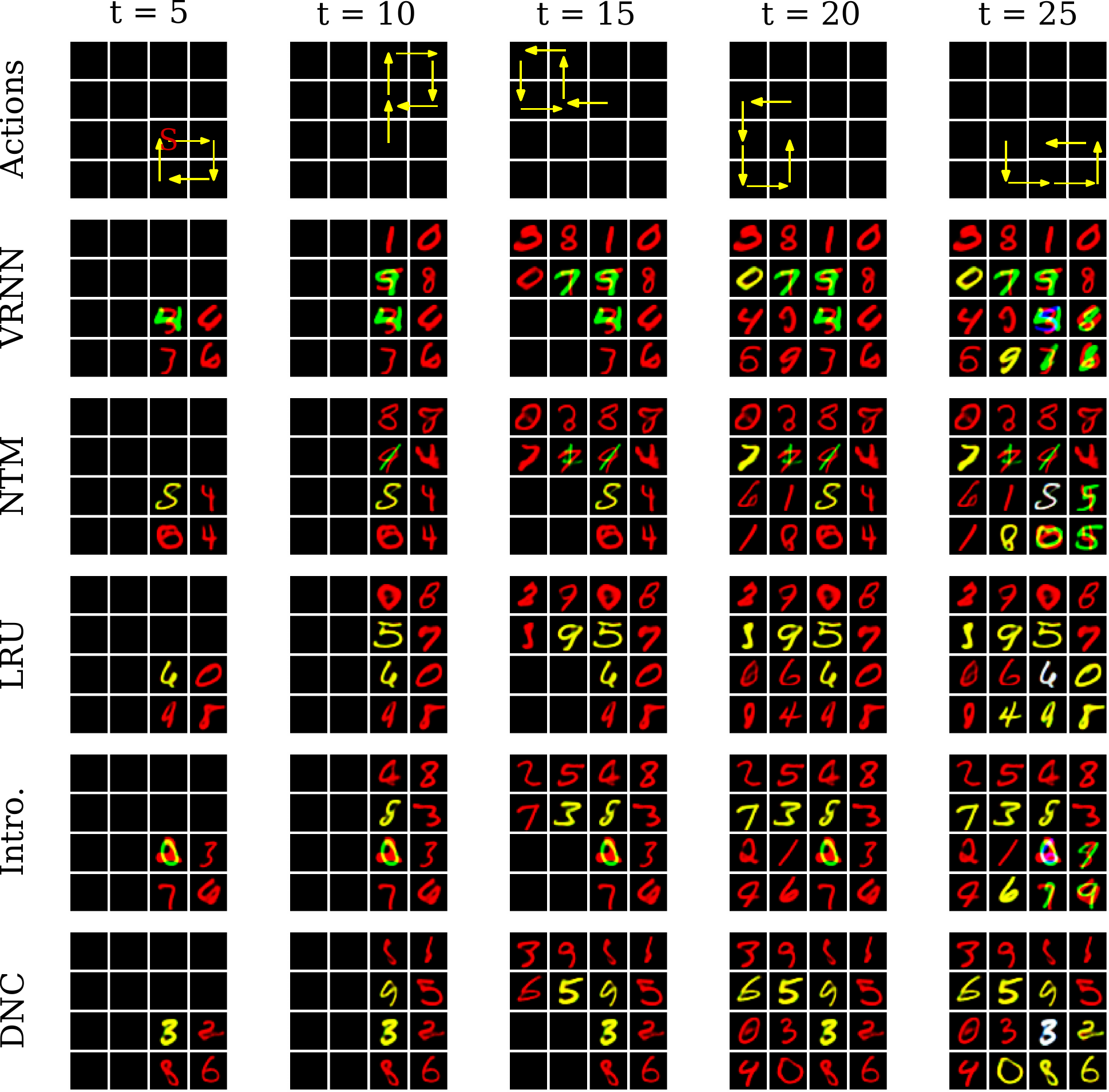}
    \caption{Generation examples in the MNIST Maze. Actions are shown in the top row. In each column of that row, the last five actions at each time point ($t=5, 10, 15, \dots$) are indicated by yellow arrows. The best four models of each type are shown in each row below. The first time a grid cell was visited, the red pixel channel was turned on; the second time, a new image was generated and super-imposed, with the red and green channels turned on (so overlaid image colour is red+green=yellow); the third time, the blue channel was turned on, so the overlaid, generated image was white. We see that the VRNN had inconsistent generations, as the subsequent visits to grid cells produced overlays of different colours that did not share the same shape. The NTM and Introspection models were by comparison better at coherent generation, and the LRU- and DNC-based GTMMs were the best here, so that, for example, the white and yellow digits masked the underlying red digit.
    \label{fig:mnist_maze_overlay}}
\end{figure}

Since the actions are generated by a random walk process, the structure of this problem is poorly captured by memory addressing mechanisms that are based on time or positional order in a sequence. Instead, models that used content-based addressing could encode the sequence of actions alongside the latent variable representations of the images. When it was necessary to predict what is present at an already visited location, content-based addressing could be used to look up the latent variables based on the action sequence. Since any grid location could be reached via multiple routes, the models had also to capture the invariance that action sequences should be converted into displacements from the origin. The DNC- and LRU-GTMM performed best at this task (Figure \ref{fig:mnist_maze_results}), exhibiting significant KL reductions as more of the environment was explored. In Fig. \ref{fig:mnist_maze_overlay}, we also show generation of a sequences in the maze by each model. Only LRU-GTMM and DNC-GTMM consistently generated the same digits when returning to the same positions.

Although the NTM-backed GTMM has content-based addressing, its ability to allocate free locations in memory is generally inferior to the abilities of models using LRUs and DNCs. The LRU- and DNC-based mechanisms could easily store new memories but also could collocate the new memories with context information registering the computed position on the grid each image was located.  

\subsection{Towards Coherent Generation in Realistic 3D Environments}
Ultimately, we wish to design agents that operate in realistic environments and can learn from sequential information, using memory to form predictions. 
These agents should possess spatio-temporally coherent memories of environments, understanding that walls typically do not shift and that undisturbing movements change camera angles but not scene arrangements. 
Our first study of this problem was to test whether GTMMs can maintain consistent predictions when provided with frames from an in-place rotation of a camera for two full turns.
These experiments used a procedurally-generated 3D maze environment with random wall configurations, textures, and object positions.

\begin{figure}[t]
	\centering
    \includegraphics[width=\linewidth]{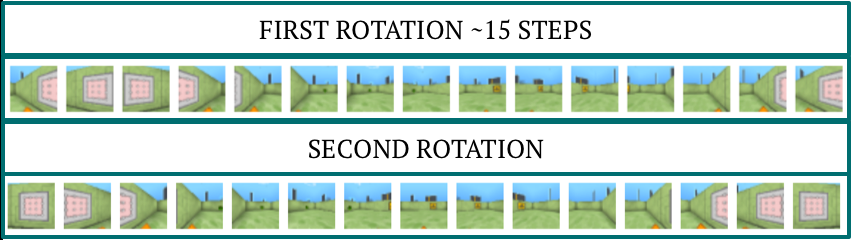}
    \caption{A training sequence for a 30 timestep in-place rotation, where the period until repetition was approximately 15 steps.}
    \label{fig:Rot15Example}
\end{figure}

The rotational dynamics of the environment included acceleration, so each frame did not represent a view from an angle that is equally distributed around the unit circle. The models had to cope with this structure. 
Additionally, because the frames were captured at discrete moments, the models had to learn to interpolate past views after the full turn, instead of merely copying frames from the first rotation.
The rotational period was $t=15$ steps, and a full episode took place over $t=30$ steps.
An example training sequence is shown in Figure \ref{fig:Rot15Example}.
A successful generative model of this data had to create random environment panoramas and generate views consistent with the panorama on the second full turn.

\begin{figure}[t]
	\centering
	\includegraphics[width=1\textwidth]{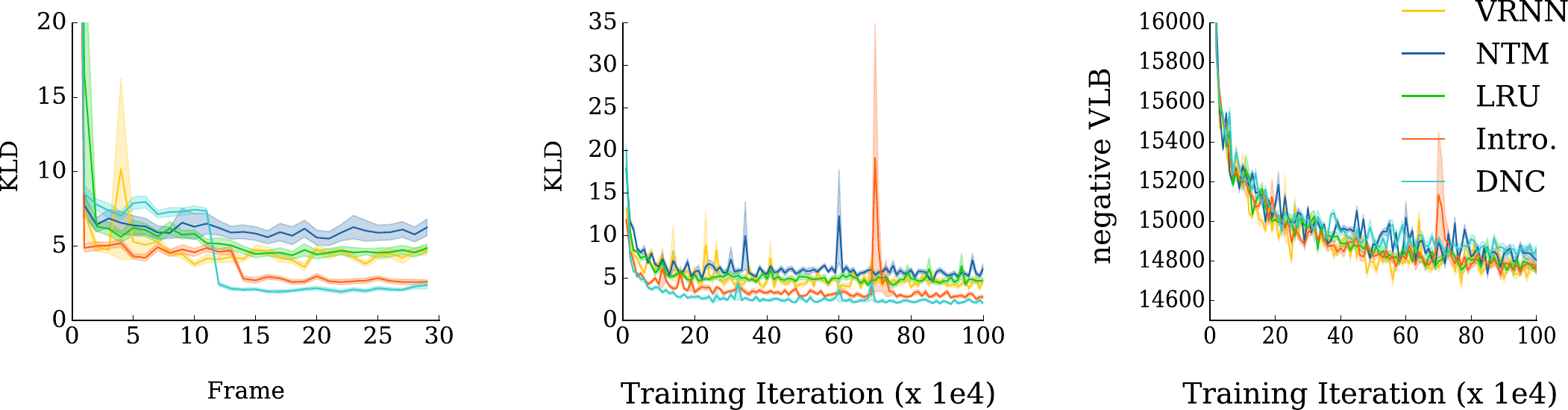}
    \caption{Realistic environments task. The VRNN had the lowest variational lower bound, but this measure tells very little of the overall story. Instead, the DNC-GTMM and Introspective-GTMM showed significant KL reductions when the sequence is predictable from memory, and this measure translates into good behaviour during generation of samples.
    \label{fig:fast_rot_summary}}
\end{figure}

In Fig. \ref{fig:fast_rot_summary}, we see that the VRNN had the lowest variational lower bound. However, in the generative samples, it is clear that the VRNN was also incoherent across time, as it forgot information about paintings on the walls and buildings on the skyline (Fig. \ref{fig:fast_rot_gen}). We argue that the more representative quantification, of more significance than the training loss, is the Kullback-Leibler divergence of the last frame, as well as the KL reduction at the time of the turn. These indicators showed that the DNC-GTMM and Introspective-GTMM models were able to predict the redundant frames from memory.

\section{Discussion and Conclusions}
\label{sect:discussion}
The aim of this paper has been to open a research direction.
We have seen that, on a range of tasks, standard generative temporal models are limited by their memory capacity and memory access mechanisms.
This has motivated the design of new generative temporal models with external memory systems whose performance is in some cases qualitatively better.

We have tried to provide proofs of concept without extraneous complication: for example, our variational distributions are simple, diagonal Gaussians; we could consider more complex posterior distributions, like those used in DRAW \citep{gregor2015draw}, normalising flows \citep{rezende2015variational}, auxiliary variables \citep{ranganath2016hierarchical}, or models with discrete variables \citep{eslami2016attend}.
We have also aimed to explore the advantages and disadvantages of a variety of external memory mechanisms for generative temporal modelling.
Our results suggest that none of the architectures we report on is uniformly dominant across tasks.
However, an interesting direction of further research is strongly suggested by our results.
Namely, we imagine that a new model, combining the direct storage of latent variables, as in the Introspective-GTMM, with content-based addressing, as in the NTM-, LRU-, and DNC-GTMM, could indeed prove to have performance that is uniformly dominant across all tasks.
Furthermore, many of these sparse memory access models can be more efficiently implemented by using fast K-nearest neighbour lookup, though we did not explore such savings here.
We leave the development of these more sophisticated models to future work.

The storage of latent variables or transformed latent variables in memory additionally suggests several intriguing extensions of our framework. 
Currently, the models are based on the mathematics of optimal filtering: they produce a sample at every time step that is drawn from the filtering posterior $q_\phi(\vz_{t}|\vz_\st, \vx_\seqt)$. 
Once latent variables are stored in memory, the formal distinction between filtering and smoothing (with a non-temporally causal posterior $q_\phi(\vz_{t}|\vx_\seqT)$) begins to break down. 
We can imagine further mechanisms that modify previously written latent variables after the fact by overwriting previously written locations. 
The explicit storage of latent variables in memory supports this, whereas it would be difficult to precisely modify the component of the latent state that encodes a historical latent variable in a more conventional, densely connected RNN.

\pagebreak

\bibliographystyle{abbrvnat}
\bibliography{references}

\begin{thebibliography}{43}
\providecommand{\natexlab}[1]{#1}
\providecommand{\url}[1]{\texttt{#1}}
\expandafter\ifx\csname urlstyle\endcsname\relax
  \providecommand{\doi}[1]{doi: #1}\else
  \providecommand{\doi}{doi: \begingroup \urlstyle{rm}\Url}\fi

\bibitem[Assael et~al.(2015)Assael, Wahlstr{\"o}m, Sch{\"o}n, and
  Deisenroth]{assael2015data}
J.-A.~M. Assael, N.~Wahlstr{\"o}m, T.~B. Sch{\"o}n, and M.~P. Deisenroth.
\newblock Data-efficient learning of feedback policies from image pixels using
  deep dynamical models.
\newblock \emph{arXiv preprint arXiv:1510.02173}, 2015.

\bibitem[Bahdanau et~al.(2014)Bahdanau, Cho, and Bengio]{bahdanau2014neural}
D.~Bahdanau, K.~Cho, and Y.~Bengio.
\newblock Neural machine translation by jointly learning to align and
  translate.
\newblock \emph{arXiv preprint arXiv:1409.0473}, 2014.

\bibitem[Bayer and Osendorfer(2014)]{bayer2014learning}
J.~Bayer and C.~Osendorfer.
\newblock Learning stochastic recurrent networks.
\newblock \emph{arXiv preprint arXiv:1411.7610}, 2014.

\bibitem[Bialek et~al.(2001)Bialek, Nemenman, and
  Tishby]{bialek2001predictability}
W.~Bialek, I.~Nemenman, and N.~Tishby.
\newblock Predictability, complexity, and learning.
\newblock \emph{Neural computation}, 13\penalty0 (11):\penalty0 2409--2463,
  2001.

\bibitem[Chung et~al.(2015{\natexlab{a}})Chung, Gulcehre, Cho, and
  Bengio]{chung2015gated}
J.~Chung, C.~Gulcehre, K.~Cho, and Y.~Bengio.
\newblock Gated feedback recurrent neural networks.
\newblock \emph{arXiv preprint arXiv:1502.02367}, 2015{\natexlab{a}}.

\bibitem[Chung et~al.(2015{\natexlab{b}})Chung, Kastner, Dinh, Goel, Courville,
  and Bengio]{chung2015recurrent}
J.~Chung, K.~Kastner, L.~Dinh, K.~Goel, A.~C. Courville, and Y.~Bengio.
\newblock A recurrent latent variable model for sequential data.
\newblock In \emph{Advances in neural information processing systems}, pages
  2962--2970, 2015{\natexlab{b}}.

\bibitem[Deisenroth and Rasmussen(2011)]{deisenroth2011pilco}
M.~Deisenroth and C.~E. Rasmussen.
\newblock Pilco: A model-based and data-efficient approach to policy search.
\newblock In \emph{Proceedings of the 28th International Conference on machine
  learning (ICML-11)}, pages 465--472, 2011.

\bibitem[Eslami et~al.(2016)Eslami, Heess, Weber, Tassa, Szepesvari, Hinton,
  et~al.]{eslami2016attend}
S.~A. Eslami, N.~Heess, T.~Weber, Y.~Tassa, D.~Szepesvari, G.~E. Hinton, et~al.
\newblock Attend, infer, repeat: Fast scene understanding with generative
  models.
\newblock In \emph{Advances In Neural Information Processing Systems}, pages
  3225--3233, 2016.

\bibitem[Fraccaro et~al.(2016)Fraccaro, S{\o}nderby, Paquet, and
  Winther]{fraccaro2016sequential}
M.~Fraccaro, S.~K. S{\o}nderby, U.~Paquet, and O.~Winther.
\newblock Sequential neural models with stochastic layers.
\newblock In \emph{Advances in Neural Information Processing Systems}, 2016.

\bibitem[Fu(2005)]{fu2005stochastic}
M.~C. Fu.
\newblock Stochastic gradient estimation.
\newblock Technical report, DTIC Document, 2005.

\bibitem[Ghahramani and Hinton(1996)]{ghahramani1996parameter}
Z.~Ghahramani and G.~E. Hinton.
\newblock Parameter estimation for linear dynamical systems.
\newblock Technical report, Technical Report CRG-TR-96-2, University of
  Totronto, Dept. of Computer Science, 1996.

\bibitem[Graves et~al.(2014)Graves, Wayne, and Danihelka]{graves2014neural}
A.~Graves, G.~Wayne, and I.~Danihelka.
\newblock Neural turing machines.
\newblock \emph{arXiv preprint arXiv:1410.5401}, 2014.

\bibitem[Graves et~al.(2016)Graves, Wayne, Reynolds, Harley, Danihelka,
  Grabska-Barwi{\'n}ska, Colmenarejo, Grefenstette, Ramalho, Agapiou,
  et~al.]{graves2016hybrid}
A.~Graves, G.~Wayne, M.~Reynolds, T.~Harley, I.~Danihelka,
  A.~Grabska-Barwi{\'n}ska, S.~G. Colmenarejo, E.~Grefenstette, T.~Ramalho,
  J.~Agapiou, et~al.
\newblock Hybrid computing using a neural network with dynamic external memory.
\newblock \emph{Nature}, 538\penalty0 (7626):\penalty0 471--476, 2016.

\bibitem[Grefenstette et~al.(2015)Grefenstette, Hermann, Suleyman, and
  Blunsom]{grefenstette2015learning}
E.~Grefenstette, K.~M. Hermann, M.~Suleyman, and P.~Blunsom.
\newblock Learning to transduce with unbounded memory.
\newblock In \emph{Advances in Neural Information Processing Systems}, pages
  1828--1836, 2015.

\bibitem[Gregor et~al.(2015)Gregor, Danihelka, Graves, Jimenez~Rezende, and
  Wierstra]{gregor2015draw}
K.~Gregor, I.~Danihelka, A.~Graves, D.~Jimenez~Rezende, and D.~Wierstra.
\newblock Draw: A recurrent neural network for image generation.
\newblock In \emph{ICML}, 2015.

\bibitem[He et~al.(2015)He, Zhang, Ren, and Sun]{he2015deep}
K.~He, X.~Zhang, S.~Ren, and J.~Sun.
\newblock Deep residual learning for image recognition.
\newblock \emph{arXiv preprint arXiv:1512.03385}, 2015.

\bibitem[Hermann et~al.(2015)Hermann, Kocisky, Grefenstette, Espeholt, Kay,
  Suleyman, and Blunsom]{hermann2015teaching}
K.~M. Hermann, T.~Kocisky, E.~Grefenstette, L.~Espeholt, W.~Kay, M.~Suleyman,
  and P.~Blunsom.
\newblock Teaching machines to read and comprehend.
\newblock In \emph{Advances in Neural Information Processing Systems}, pages
  1693--1701, 2015.

\bibitem[Hochreiter and Schmidhuber(1997)]{hochreiter1997long}
S.~Hochreiter and J.~Schmidhuber.
\newblock Long short-term memory.
\newblock \emph{Neural computation}, 9\penalty0 (8):\penalty0 1735--1780, 1997.

\bibitem[Joulin and Mikolov(2015)]{joulin2015inferring}
A.~Joulin and T.~Mikolov.
\newblock Inferring algorithmic patterns with stack-augmented recurrent nets.
\newblock In \emph{Advances in Neural Information Processing Systems}, pages
  190--198, 2015.

\bibitem[Kadlec et~al.(2016)Kadlec, Schmid, Bajgar, and
  Kleindienst]{kadlec2016text}
R.~Kadlec, M.~Schmid, O.~Bajgar, and J.~Kleindienst.
\newblock Text understanding with the attention sum reader network.
\newblock \emph{arXiv preprint arXiv:1603.01547}, 2016.

\bibitem[Kalman(1960)]{kalman1960new}
R.~E. Kalman.
\newblock A new approach to linear filtering and prediction problems.
\newblock \emph{Journal of basic Engineering}, 82\penalty0 (1):\penalty0
  35--45, 1960.

\bibitem[Kingma and Ba(2014)]{CorrKingma2014}
D.~P. Kingma and J.~Ba.
\newblock Adam: A method for stochastic optimization.
\newblock \emph{CoRR}, 2014.

\bibitem[Kingma and Welling(2014)]{kingma2014stochastic}
D.~P. Kingma and M.~Welling.
\newblock Auto-encoding variational {B}ayes.
\newblock In \emph{ICLR}, 2014.

\bibitem[Krishnan et~al.(2015)Krishnan, Shalit, and Sontag]{krishnan2015deep}
R.~G. Krishnan, U.~Shalit, and D.~Sontag.
\newblock Deep kalman filters.
\newblock \emph{arXiv preprint arXiv:1511.05121}, 2015.

\bibitem[Kumar et~al.(2015)Kumar, Irsoy, Su, Bradbury, English, Pierce,
  Ondruska, Gulrajani, and Socher]{kumar2015ask}
A.~Kumar, O.~Irsoy, J.~Su, J.~Bradbury, R.~English, B.~Pierce, P.~Ondruska,
  I.~Gulrajani, and R.~Socher.
\newblock Ask me anything: Dynamic memory networks for natural language
  processing.
\newblock \emph{arXiv preprint arXiv:1506.07285}, 2015.

\bibitem[Lake et~al.(2015)Lake, Salakhutdinov, and Tenenbaum]{lake2015human}
B.~M. Lake, R.~Salakhutdinov, and J.~B. Tenenbaum.
\newblock Human-level concept learning through probabilistic program induction.
\newblock \emph{Science}, 350\penalty0 (6266):\penalty0 1332--1338, 2015.

\bibitem[Levine and Abbeel(2014)]{levine2014learning}
S.~Levine and P.~Abbeel.
\newblock Learning neural network policies with guided policy search under
  unknown dynamics.
\newblock In \emph{Advances in Neural Information Processing Systems}, pages
  1071--1079, 2014.

\bibitem[Li et~al.(2016)Li, Zhu, and Zhang]{li2016learning}
C.~Li, J.~Zhu, and B.~Zhang.
\newblock Learning to generate with memory.
\newblock \emph{arXiv preprint arXiv:1602.07416}, 2016.

\bibitem[Pearlmutter(1995)]{pearlmutter1995gradient}
B.~A. Pearlmutter.
\newblock Gradient calculations for dynamic recurrent neural networks: A
  survey.
\newblock \emph{IEEE Transactions on Neural networks}, 6\penalty0 (5):\penalty0
  1212--1228, 1995.

\bibitem[Rabiner(1989)]{rabiner1989tutorial}
L.~R. Rabiner.
\newblock A tutorial on hidden markov models and selected applications in
  speech recognition.
\newblock \emph{Proceedings of the IEEE}, 77\penalty0 (2):\penalty0 257--286,
  1989.

\bibitem[Ranganath et~al.(2016)Ranganath, Tran, and
  Blei]{ranganath2016hierarchical}
R.~Ranganath, D.~Tran, and D.~M. Blei.
\newblock Hierarchical variational models.
\newblock In \emph{International Conference on Machine Learning}, 2016.

\bibitem[Reed and de~Freitas(2015)]{reed2015neural}
S.~Reed and N.~de~Freitas.
\newblock Neural programmer-interpreters.
\newblock \emph{arXiv preprint arXiv:1511.06279}, 2015.

\bibitem[Rezende and Mohamed(2015)]{rezende2015variational}
D.~J. Rezende and S.~Mohamed.
\newblock Variational inference with normalizing flows.
\newblock \emph{arXiv preprint arXiv:1505.05770}, 2015.

\bibitem[Rezende et~al.(2014)Rezende, Mohamed, and
  Wierstra]{rezende2014stochastic}
D.~J. Rezende, S.~Mohamed, and D.~Wierstra.
\newblock Stochastic backpropagation and approximate inference in deep
  generative models.
\newblock In \emph{ICML}, 2014.

\bibitem[Riedel et~al.(2016)Riedel, Bo{\v{s}}njak, and
  Rockt{\"a}schel]{riedel2016programming}
S.~Riedel, M.~Bo{\v{s}}njak, and T.~Rockt{\"a}schel.
\newblock Programming with a differentiable forth interpreter.
\newblock \emph{arXiv preprint arXiv:1605.06640}, 2016.

\bibitem[Santoro et~al.(2016)Santoro, Bartunov, Botvinick, Wierstra, and
  Lillicrap]{santoro2016one}
A.~Santoro, S.~Bartunov, M.~Botvinick, D.~Wierstra, and T.~Lillicrap.
\newblock One-shot learning with memory-augmented neural networks.
\newblock \emph{arXiv preprint arXiv:1605.06065}, 2016.

\bibitem[S{\"a}rkk{\"a}(2013)]{sarkka2013bayesian}
S.~S{\"a}rkk{\"a}.
\newblock \emph{Bayesian filtering and smoothing}, volume~3.
\newblock Cambridge University Press, 2013.

\bibitem[Sukhbaatar et~al.(2015)Sukhbaatar, Weston, Fergus,
  et~al.]{sukhbaatar2015end}
S.~Sukhbaatar, J.~Weston, R.~Fergus, et~al.
\newblock End-to-end memory networks.
\newblock In \emph{Advances in neural information processing systems}, pages
  2440--2448, 2015.

\bibitem[Sutton(1991)]{sutton1991dyna}
R.~S. Sutton.
\newblock Dyna, an integrated architecture for learning, planning, and
  reacting.
\newblock \emph{ACM SIGART Bulletin}, 2\penalty0 (4):\penalty0 160--163, 1991.

\bibitem[Tornio et~al.(2007)Tornio, Honkela, and Karhunen]{tornio2007time}
M.~Tornio, A.~Honkela, and J.~Karhunen.
\newblock Time series prediction with variational bayesian nonlinear
  state-space models.
\newblock In \emph{Proc. European Symp. on Time Series Prediction
  (ESTSP’07)}, pages 11--19, 2007.

\bibitem[Vinyals et~al.(2015)Vinyals, Fortunato, and
  Jaitly]{vinyals2015pointer}
O.~Vinyals, M.~Fortunato, and N.~Jaitly.
\newblock Pointer networks.
\newblock In \emph{Advances in Neural Information Processing Systems}, pages
  2692--2700, 2015.

\bibitem[Watter et~al.(2015)Watter, Springenberg, Boedecker, and
  Riedmiller]{watter2015embed}
M.~Watter, J.~Springenberg, J.~Boedecker, and M.~Riedmiller.
\newblock Embed to control: A locally linear latent dynamics model for control
  from raw images.
\newblock In \emph{Advances in Neural Information Processing Systems}, pages
  2746--2754, 2015.

\bibitem[Weston et~al.(2014)Weston, Chopra, and Bordes]{weston2014memory}
J.~Weston, S.~Chopra, and A.~Bordes.
\newblock Memory networks.
\newblock \emph{arXiv preprint arXiv:1410.3916}, 2014.

\end{thebibliography}
\vfill

\appendix
\section{Per Step Variational Lower Bound}
\label{appdx:seqVIeqns}
The variational bound gives
\begin{align}
\log p(\vx) & = \log \int d \vz \, p(\vx , \vz) \\
& = \log \int d \vz \, p(\vx, \vz) \frac{q(\vz | \vx)}{q(\vz | \vx)} \\
& = \log \E_{q(\vz | \vx)} \frac{p(\vx, \vz)}{q(\vz | \vx)} \\
& \geq \E_{q(\vz | \vx)} \log \frac{p(\vx, \vz)}{q(\vz | \vx)} \, \, \, \, \text{(by Jensen's ineq.)}.
\end{align}
For sequences of random variables, the corresponding inequality is
\begin{align}
\log p(\vx_{\leq T}) & \geq \E_{q(\vz_{\leq T}| \vx_{\leq T})} \log \frac{p(\vx_{\leq T}, \vz_{\leq T})}{q(\vz_{\leq T} | \vx_{\leq T})}. \label{eq:vlb_series}
\end{align}
In our formulation, we assume the factorisations 
\begin{equation*}
p(\vx_{\leq T}, \vz_{\leq T}) = \prod_{t=1}^T p(\vx_t | \vz_{\leq T}, \vx_{<t}) p(\vz_t | \vz_{<t}, \vx_{<t})
\end{equation*}
and 
\begin{equation*}
q(\vz_{\leq T} | \vx_{\leq T}) = \prod_{t=1}^T q(\vz_{t} | \vz_{<t}, \vx_{\leq t}).
\end{equation*}

These allow us to convert equation \ref{eq:vlb_series} into a per time-step bound. We have:
\begin{align}
 \log p(\vx_{\leq T}) & \ge \E_{\prod_{t=1}^T q(\mathbf{z}_{t}|\mathbf{z}_{<t}, \mathbf{x}_{\le t})} \left[ \sum_{t=1}^T \text{log }p(\mathbf{x}_t| \mathbf{z}_{\le t}, \mathbf{x}_{<t}) + \text{log } p(\mathbf{z}_t| \mathbf{z}_{<t}, \mathbf{x}_{<t})
- \text{log } q(\mathbf{z}_{t}|\mathbf{z}_{<t}, \mathbf{x}_{\le t})\right] \\
 & = \E_{\prod_{t=1}^T q(\mathbf{z}_{t}|\mathbf{z}_{<t}, \mathbf{x}_{\le t})} \left[ \sum_{t=1}^T \log p(\mathbf{x}_t| \mathbf{z}_{\le t}, \mathbf{x}_{<t}) + \log \frac{p(\mathbf{z}_t| \mathbf{z}_{<t}, \mathbf{x}_{<t})}{ q(\mathbf{z}_{t}|\mathbf{z}_{<t}, \mathbf{x}_{\le t})}\right].
\end{align} 

Let $C_t \equiv \log p(\mathbf{x}_t| \mathbf{z}_{\le t}, \mathbf{x}_{<t}) + \log \frac{p(\mathbf{z}_t| \mathbf{z}_{<t}, \mathbf{x}_{<t})}{ q(\mathbf{z}_{t}|\mathbf{z}_{<t}, \mathbf{x}_{\le t})}$. Then:
\begin{align}
 \log p(\vx_{\leq T}) & \geq \E_{\prod_{t=1}^T q(\mathbf{z}_{t}|\mathbf{z}_{<t}, \mathbf{x}_{\le t})}
 \left[ \sum_{t=1}^T C_t \right] \\
 & =
 \int_{z_1}q(\mathbf{z}_{1}|\mathbf{x}_{1})
 \int_{z_2}q(\mathbf{z}_{2}|\mathbf{z}_{1}, \mathbf{x}_{\le 2})
 \dots
 \int_{z_T}q(\mathbf{z}_{T}|\mathbf{z}_{<T}, \mathbf{x}_{\le T}) \sum_{t=1}^T C_t.
\end{align} 
Each $C_t$ is not a function of the elements of the set $\{z_{t+1}, z_{t+2}, \dots, z_T\}$. Therefore, we can move each $C_t$ out from the integrals involving only those terms:
\begin{align}
 & =
 \int_{z_1}q(\mathbf{z}_{1}|\mathbf{x}_{1}) C_1
 \int_{z_2}q(\mathbf{z}_{2}|\mathbf{z}_{1}, \mathbf{x}_{\le 2})
 \dots
 \int_{z_T}q(\mathbf{z}_{T}|\mathbf{z}_{<T}, \mathbf{x}_{\le T}) \nonumber \\
 & + \int_{z_1}q(\mathbf{z}_{1}|\mathbf{x}_{1})
 \int_{z_2}q(\mathbf{z}_{2}|\mathbf{z}_{1}, \mathbf{x}_{\le 2}) C_2
 \dots \int_{z_T}q(\mathbf{z}_{T}|\mathbf{z}_{<T}, \mathbf{x}_{\le T}) \nonumber \\
 & + \dots \nonumber \\
 & + \int_{z_1}q(\mathbf{z}_{1}|\mathbf{x}_{1})
 \int_{z_2}q(\mathbf{z}_{2}|\mathbf{z}_{1}, \mathbf{x}_{\le 2})
 \dots
 \int_{z_T}q(\mathbf{z}_{T}|\mathbf{z}_{<T}, \mathbf{x}_{\le T}) C_T
\end{align} 
The interior integrals all sum to 1 since the $q$-s are distributions. Thus, we have:
\begin{align}
 & =
 \int_{z_1}q(\mathbf{z}_{1}|\mathbf{x}_{1}) C_1 \nonumber \\
 & + \int_{z_1}q(\mathbf{z}_{1}|\mathbf{x}_{1})
 \int_{z_2}q(\mathbf{z}_{2}|\mathbf{z}_{1}, \mathbf{x}_{\le 2}) C_2 \nonumber \\
 & + \dots \nonumber \\
 & + \int_{z_1}q(\mathbf{z}_{1}|\mathbf{x}_{1})
 \int_{z_2}q(\mathbf{z}_{2}|\mathbf{z}_{1}, \mathbf{x}_{\le 2})
 \dots
 \int_{z_T}q(\mathbf{z}_{T}|\mathbf{z}_{<T}, \mathbf{x}_{\le T}) C_T. 
\end{align}
We can write this more simply as
\begin{equation}
 \mathcal{F} =
 \sum^T_{t=1}
 \E_{\prod_{\tau=1}^{t} q(\mathbf{z}_{\tau}|\mathbf{z}_{<\tau}, \mathbf{x}_{\le \tau})} C_t.
\end{equation}
Finally, we bring this expression into a more conventional form by writing
\begin{align}
 \mathcal{F} & =
 \sum^T_{t=1}
 \E_{\prod_{\tau=1}^{t} q(\mathbf{z}_{\tau}|\mathbf{z}_{<\tau}, \mathbf{x}_{\le \tau})} \bigg[ \log p(\mathbf{x}_t| \mathbf{z}_{\le t}, \mathbf{x}_{<t}) + \log \frac{p(\mathbf{z}_t| \mathbf{z}_{<t}, \mathbf{x}_{<t})}{ q(\mathbf{z}_{t}|\mathbf{z}_{<t}, \mathbf{x}_{\le t})} \bigg] \nonumber  \\
& =
 \sum^T_{t=1}
 \E_{\prod_{\tau=1}^{t-1}  q(\mathbf{z}_{\tau}|\mathbf{z}_{<\tau}, \mathbf{x}_{\le \tau})} \bigg[ \E_{q(\mathbf{z}_{t}|\mathbf{z}_{<t}, \mathbf{x}_{\le t})} \bigg( \log p(\mathbf{x}_t| \mathbf{z}_{\le t}, \mathbf{x}_{<t}) + \log \frac{p(\mathbf{z}_t| \mathbf{z}_{<t}, \mathbf{x}_{<t})}{ q(\mathbf{z}_{t}|\mathbf{z}_{<t}, \mathbf{x}_{\le t})} \bigg) \bigg] \nonumber  \\
& =
 \sum^T_{t=1}
 \E_{\prod_{\tau=1}^{t-1}  q(\mathbf{z}_{\tau}|\mathbf{z}_{<\tau}, \mathbf{x}_{\le \tau})} \bigg[ \E_{q(\mathbf{z}_{t}|\mathbf{z}_{<t}, \mathbf{x}_{\le t})} \log p(\mathbf{x}_t| \mathbf{z}_{\le t}, \mathbf{x}_{<t}) \nonumber \\
 & \hspace{5cm} - \text{KL}[q(\mathbf{z}_{t}|\mathbf{z}_{<t}, \mathbf{x}_{\le t}) || p(\mathbf{z}_t| \mathbf{z}_{<t}, \mathbf{x}_{<t})] \bigg]. 
\end{align}
Bringing back the distributional parameters $p_\theta$ and $q_\phi$ yields an equation equivalent to the main text.

\section{Visual Architectures}
\label{appen:visualArchitectures}
\subsection{Tasks with Digits and Characters} 
\begin{figure}[H]
    \centering
    \includegraphics{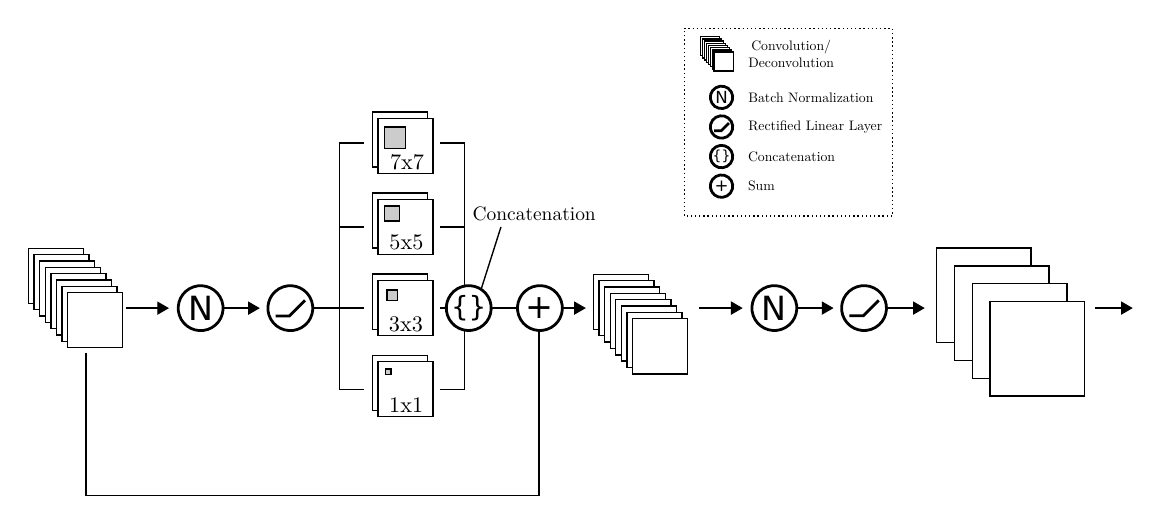}
    \caption{The first visual block.}
    \label{fig:cnn}
\end{figure}

The  posterior  map -- that is, the mapping from inputs $x_t$ to latents $z_t$ -- was implemented as a convolutional neural network (CNN). The CNN consisted of two ``blocks," arranged in series (the first block is shown in Figure \ref{fig:cnn}). The first block was fed a $1 \times 28\times 28$ grey-scale image as input (rescaling the input images if necessary), where the saturation was mapped into the range $(-1,1)$. This input was passed to four, parallel, dimension-preserving convolutional streams, which each convolved the input using $8$ kernels of size $1 \times 1$, $3 \times 3$, $5 \times 5$, and $7 \times 7$, respectively, padding as necessary for dimension-preservation. The outputs from the parallel streams were passed through a batch-normalization layer and a rectified-linear layer, before being concatenated to a total of 32 feature-maps. These 32 feature-maps served as the input to a dimension-halving convolution using a kernel of size $3 \times 3$, which was followed by a batch normalization layer and a rectified linear layer. Thus, this first processing block functioned to take in a $32 \times 32$ image as input and return 32 feature-maps of size $16 \times 16$. A second, identical block followed from the first, except for two differences: there was no batch normalization and rectified layer after the final convolution, and the final number of kernels was 64. 
To produce a sample of a latent variable, we then mapped the kernels through a linear layer to a vector of length $64$. 32 of these dimensions were used to construct the mean $\mu$ and $32$ were used to construct the vector $\log \sigma$ for the parameters of a Gaussian distribution. Together with a sample from a standard normal $\epsilon$, the latent was generated as $z_t = \mu_t + \sigma_t \epsilon_t$, thus completing the posterior map.

The observation map -- the mapping from latents and any recurrent deterministic variables to reconstructions $\hat{x}_t$ -- was identical to the posterior map, except convolution operations were replaced with deconvolution operations. 

\subsection{Tasks Involving Frames of 3D Environments}
For the three-dimensional environment visual model, our inputs included colour channels comprising $3 \times 32 \times 32$ values. 
We used two separate pathways, one roughly to encode global information across an image, and another roughly to encode local textures.
The global information pathway convolved the image using $128$ $5 \times 5$ kernels with stride 3, no padding. These were then passed through $256$ $4 \times 4$ kernels with stride 2 without padding. Then the feature-maps were convolved with $256$ $4 \times 4$ kernels with stride 1 without padding. This gave $256$ $1 \times 1$ super-pixels.

The local texture pathway had a convolutional layer with 16 kernels of size $4 \times 4$, stride 2, padding of 1. The feature-maps were passed through another convolutional layer with 8 kernels of size $4 \times 4$, stride 2, and padding of 1. This yielded a block of size $8 \times 8 \times 8$, which was linearised to $512$ vector elements and concatenated with the $256$ features from the global information pathway. 

These were passed through a linear layer to produce a 250-dimensional $\mu$ and 250-dimensional $\log \sigma$ for the latent distribution.

The observation map was the transpose of this model as before.

\section{Generated Samples for Select Tasks}
\label{appen:gen_samples}
\begin{figure}[H]
	\centering
	\includegraphics[width=1\textwidth]{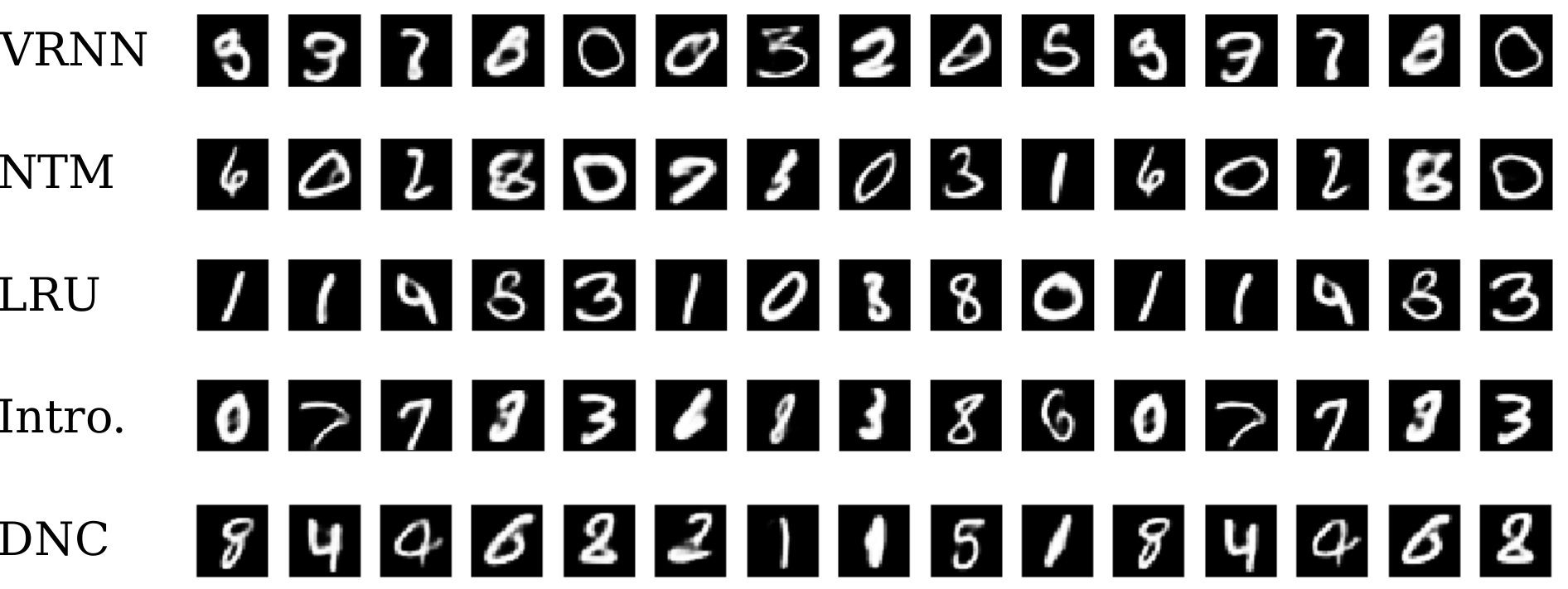}
    \caption{Perfect recall with $l=10$, $k=5$. The last $k=5$ frames should match the first 5.}
\end{figure}

\begin{figure}[H]
	\centering
	\includegraphics[width=1\textwidth]{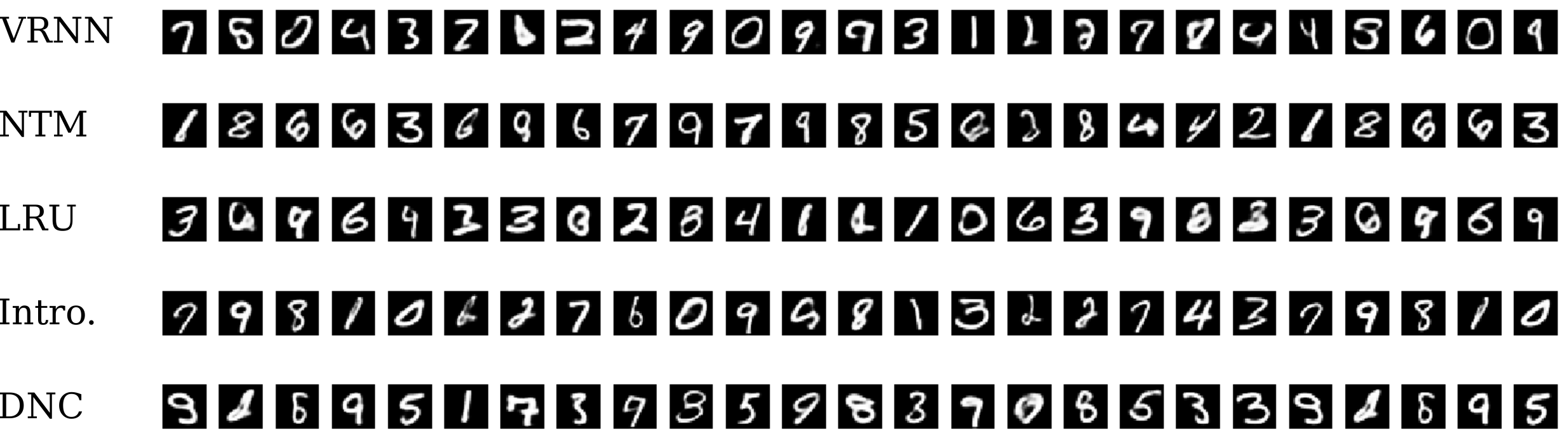}
    \caption{Perfect recall with $l=20$, $k=5$. The last $k=5$ frames should match the first 5.}
\end{figure}

\begin{figure}[H]
	\centering
	\includegraphics[width=1\textwidth]{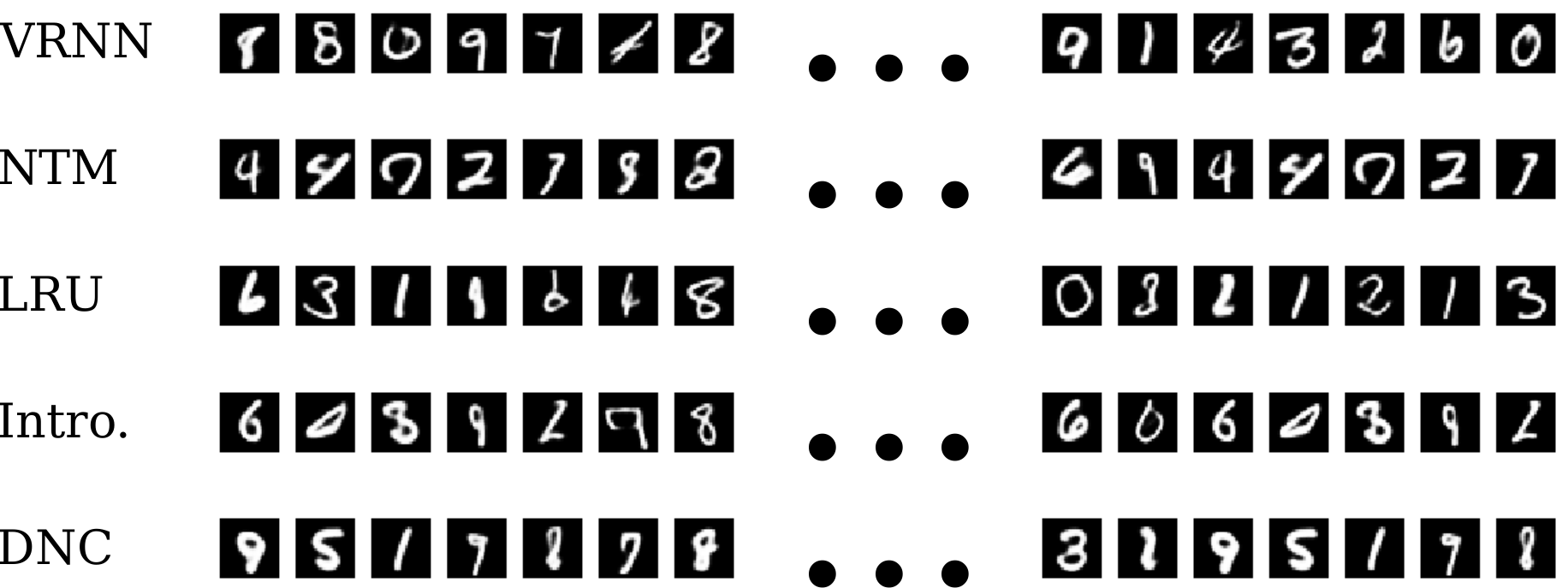}
    \caption{Perfect recall with $l=50$, $k=5$. Intermediate frames are omitted. The last $k=5$ frames should match the first 5.}
\end{figure}

\begin{figure}[H]
	\centering
	\includegraphics[width=1\textwidth]{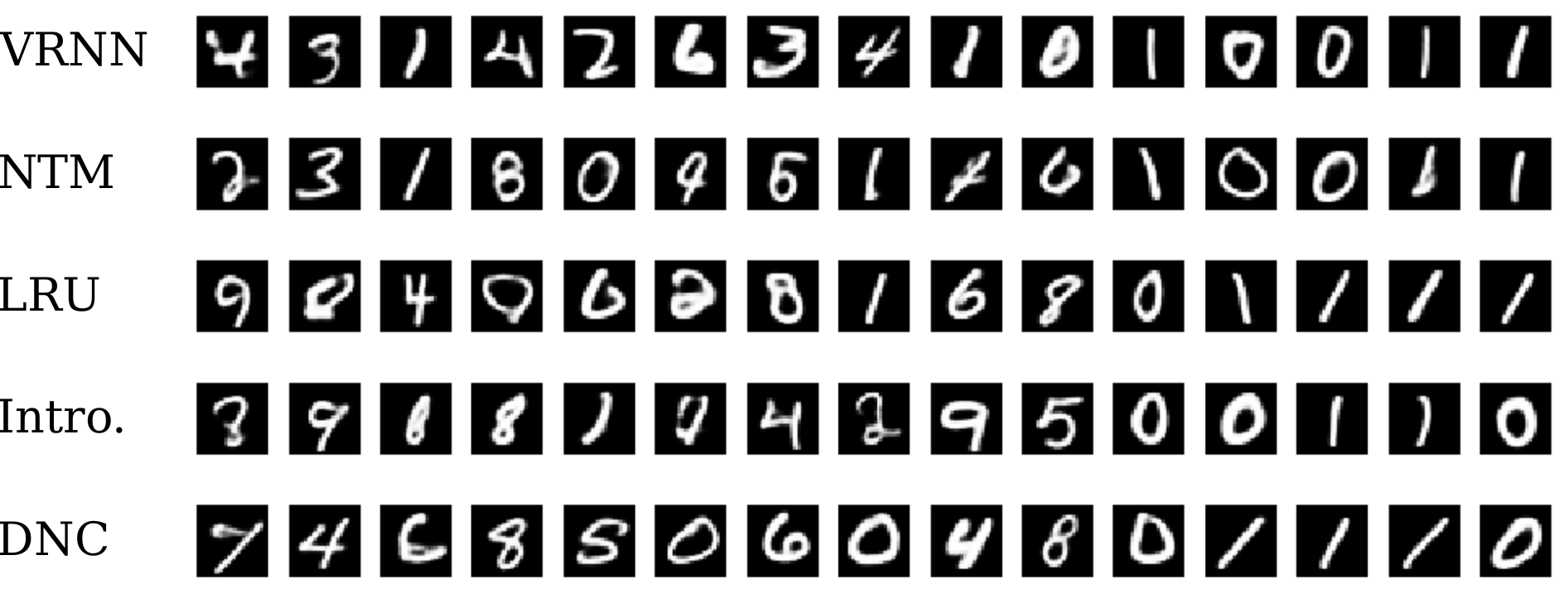}
    \caption{Parity recall with $l=10$, $k=5$. Over the last 5 frames, a generated $0$ indicates should correspond to an even digit in the first five frames, and a generated $1$ should correspond to an odd digit.}
\end{figure}

\begin{figure}[H] 
	\centering
	\includegraphics[width=1\textwidth]{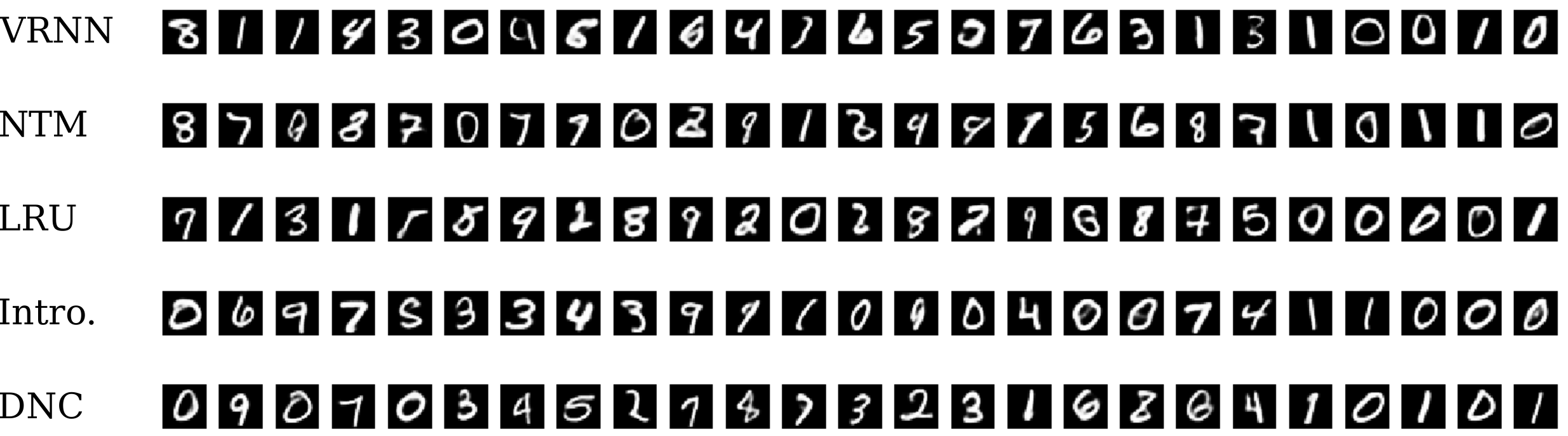}
    \caption{Parity recall with $l=20$, $k=5$. Same as above.}
\end{figure}

\begin{figure}[H]
	\centering
	\includegraphics[width=1\textwidth]{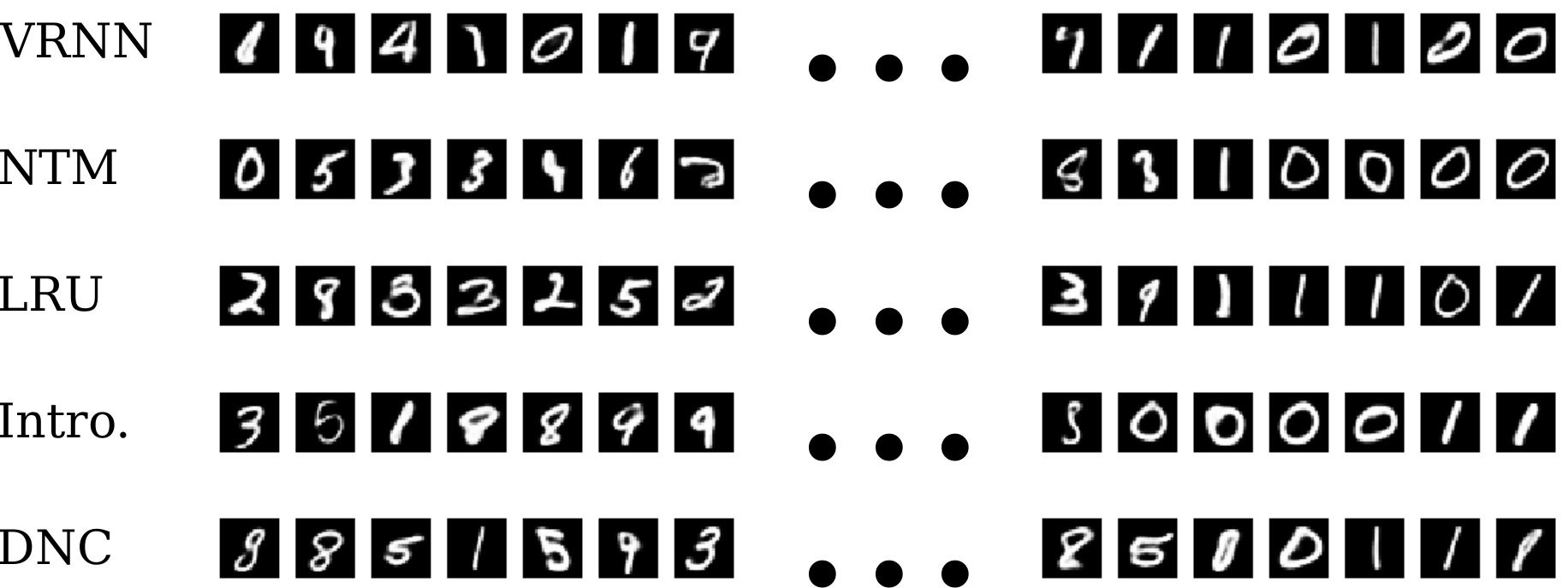}
    \caption{Parity recall with $l=50$, $k=5$.}
\end{figure}

\begin{figure}[H]
	\centering
	\includegraphics[width=1\textwidth]{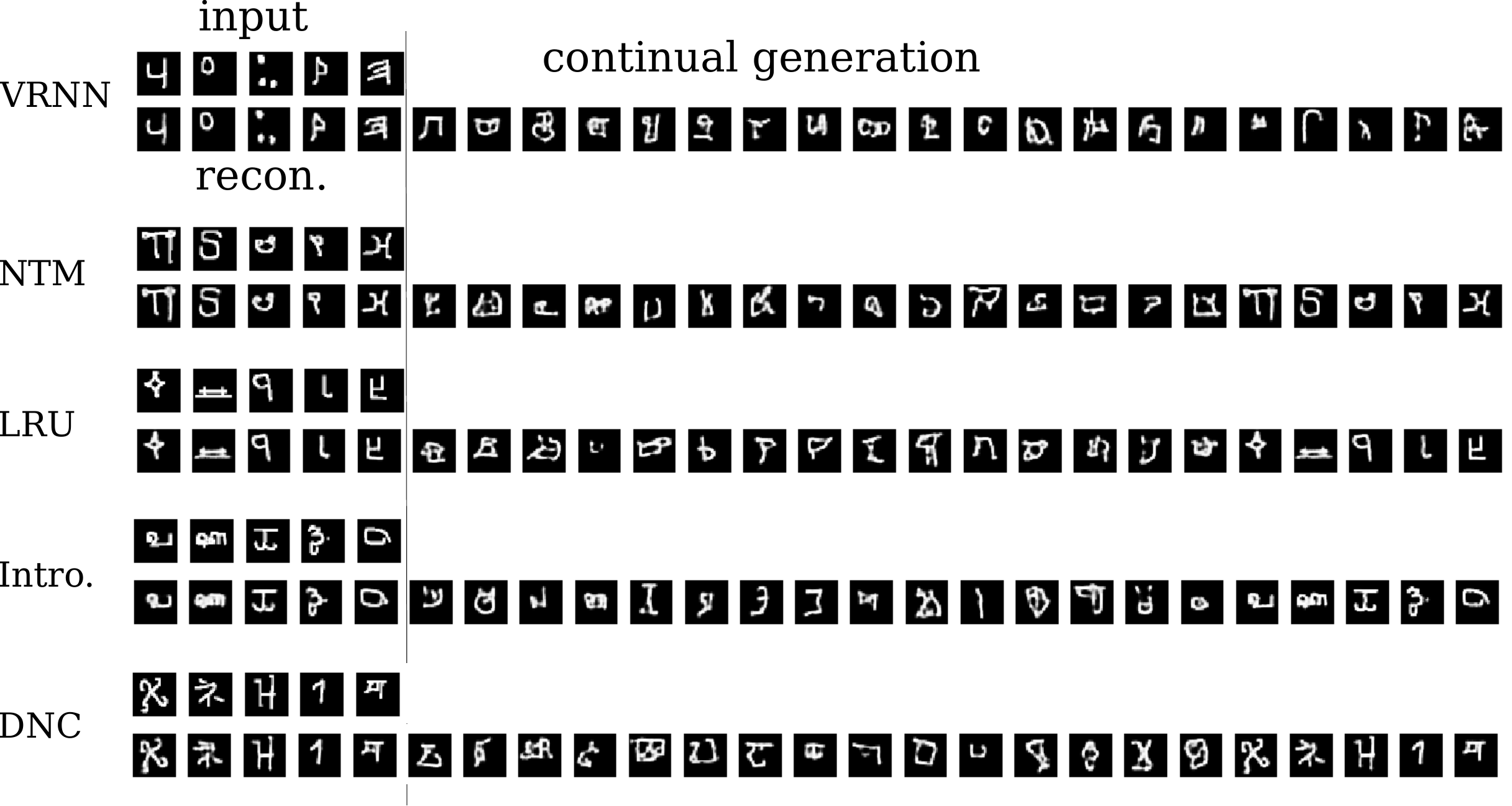}
    \caption{One-shot recall with $l=20$, $k=5$. The last 5 generated images should match the first 5, even though these images were held out from the training set.}
\end{figure}

\begin{figure}[H]
	\centering
	\includegraphics[width=1\textwidth]{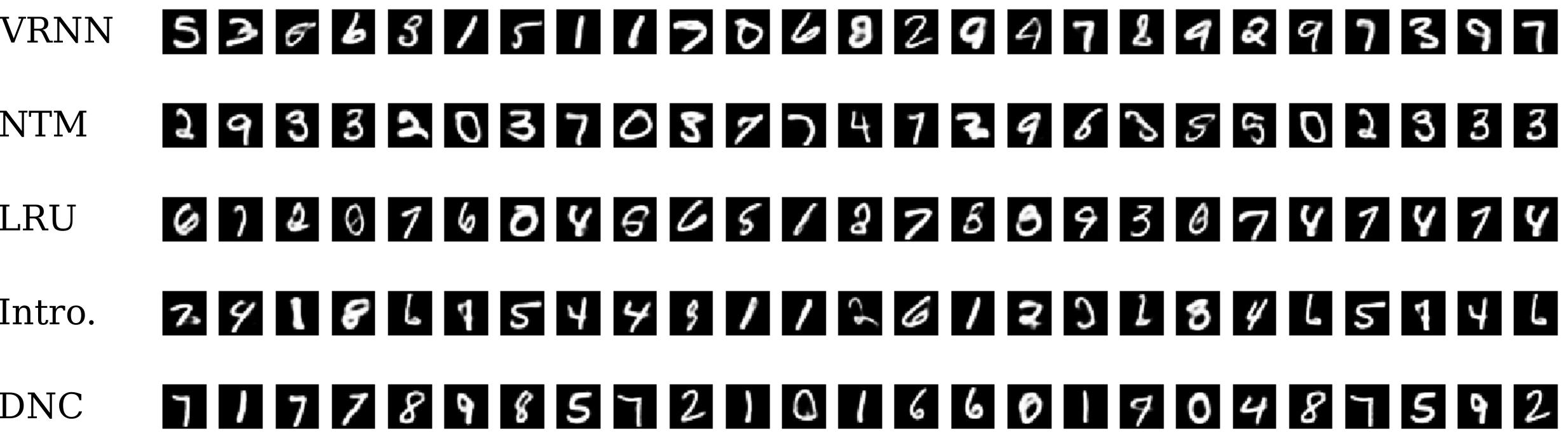}
    \caption{Dynamic dependency task with $l=20$, $k=5$. Starting from the fifth-to-last frame, the numeric value of each digit should indicate the temporal position of the digit to retrieve next.}
\end{figure}

\begin{figure}[H]
	\centering
	\includegraphics[width=1\textwidth]{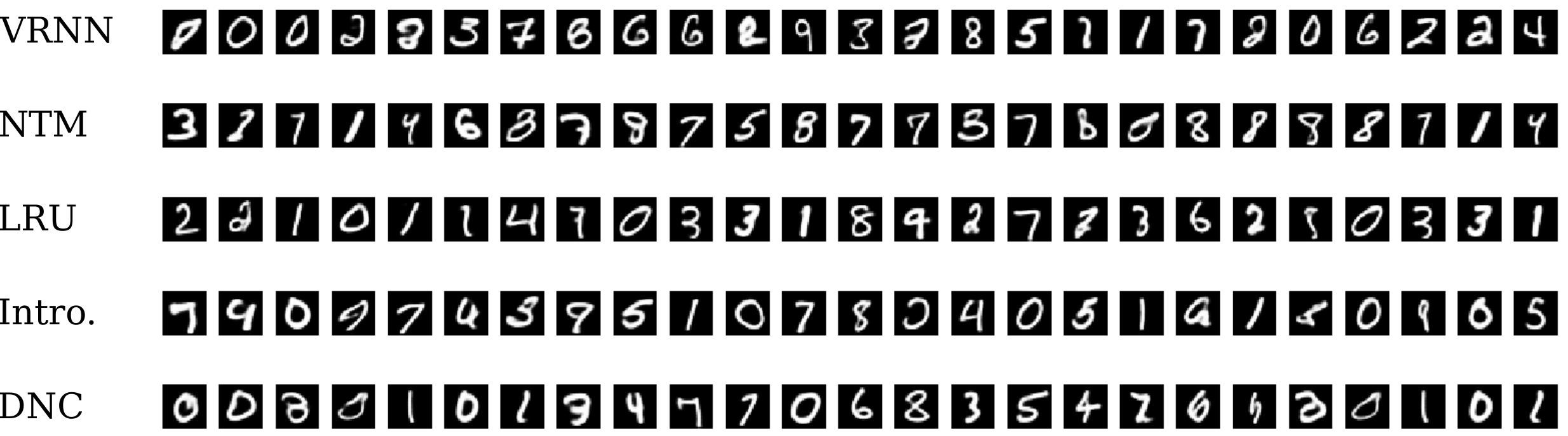}
    \caption{Content-based recall with $l=20$, $k=5$. The fifth-to-last frame is a repeat of a frame in the pre-recall phase. A working model should have continued to generate the same sequence from there.}
\end{figure}

\begin{landscape}
\begin{figure}[t]
	\centering
	\includegraphics[angle=0, width=1.6\textwidth]{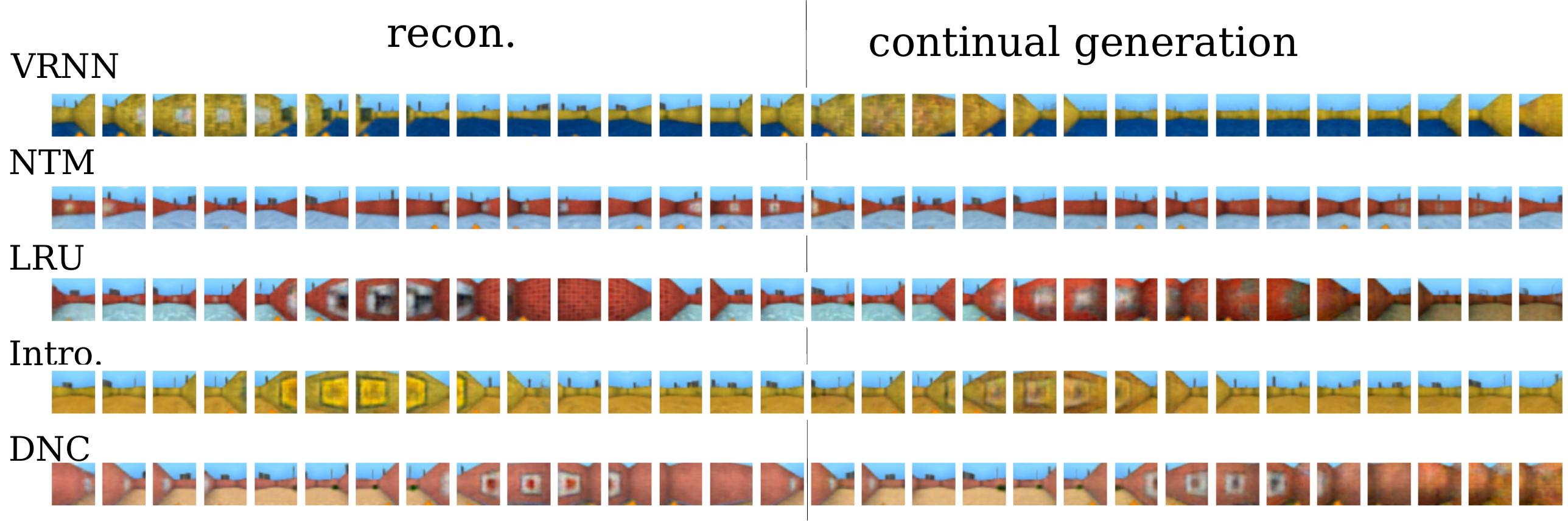}
    \caption{Rotation in a complex environment. The models observed 15 frames from the environment during one rotation, then generated corresponding frames during a second rotation. The VRNN model clearly forgot information from the first rotation, but the other models clearly demonstrated memory of paintings on the walls, floor colour, and buildings on the skyline.
    \label{fig:fast_rot_gen}}
\end{figure}
\end{landscape}
\section{Pseudocode for Generating Artificial Sequences with Characters and Digits}
\label{pseudocodeForTasks}
\begin{algorithm}[H]
  \caption{Training Sequence Generation for Perfect Recall/One-shot Recall Tasks}\label{pseudoPerfectRecall}
  \begin{algorithmic}[1]
\scriptsize{
    \Procedure{PerfectRecall/OneShotRecall}{}
      \State $\textit{generator} \gets \text{instance generator from dataset (MNIST/OmniGlot)}$
	  \State $\textit{l} \gets \text{length of random sequence}$
      \State $\textit{k} \gets \text{length of recall interval}$
      \State $\textit{sequence} \gets  \{\}$
      \\
      \Comment{Pick a random sequence of instances from the dataset}
      \For{$i = 1,$ \textit{l}}
         \State \textit{sequence}$[i]$ $\gets s$ $\sim$  \textit{generator}$()$ 
      \EndFor
      \Comment{Repeat instances from the beginning of the sequence}
      \For{$j =  1,$ \textit{k}}
        \State append(\textit{sequence}, \textit{sequence}$[j]$) 
      \EndFor
      \State \textbf{return} \textit{sequence}\Comment{Return the full sequence}
    \EndProcedure
}
  \end{algorithmic}
\end{algorithm}

\begin{algorithm}[H]
  \caption{Training Sequence Generation for Parity Recall Task}\label{pseudoParityRecall}
  \begin{algorithmic}[1]
\scriptsize{
     \Procedure{ParityRecall}{}
      \State $\textit{generator} \gets \text{instance generator from dataset (MNIST)}$
	  \State $\textit{l} \gets \text{length of random sequence}$
      \State $\textit{k} \gets \text{length of recall interval}$
      \State $\textit{sequence}, \textit{labels} \gets  \{\}, \{\}$
      \\
      \Comment{Pick a random sequence of instances from}
      
       \Comment{the dataset and their labels}
      \For{{$i = 1,$ \textit{l}}} 
         \State \textit{sequence}$[i]$, \textit{labels}$[i]$ $\gets s, l_s$ $\sim$  \textit{generator}$()$ 
      \EndFor
      \\
      \Comment{Append random $0$ or $1$ instances based on the parity of the}
      
      \Comment{labels of instances from beginning of the sequence}      
      \For{$j =  1,$ \textit{k}}
        \State append(\textit{sequence}, $s$ $\sim$\textit{generator}$(label = $ parity(\textit{labels}$[j])))$
      \EndFor
      \State \textbf{return} \textit{sequence}\Comment{Return full sequence}
    \EndProcedure
}
  \end{algorithmic}
\end{algorithm}

\begin{algorithm}[H]
  \caption{Training Sequence Generation for Dynamic Dependency Task}\label{pseudoDynamicDependency}
  \begin{algorithmic}[1]
\scriptsize{
    \Procedure{DynamicDependency}{}
      \State $\textit{generator} \gets \text{instance generator from dataset (MNIST)}$
	  \State $\textit{l} \gets \text{length of random sequence}$
      \State $\textit{k} \gets \text{length of recall interval}$
      \State $\textit{sequence}, \textit{labels} \gets  \{\}, \{\}$
      \\
      \Comment{Pick a random sequence of instances from}
      
       \Comment{the dataset and their labels}
      \For{{$i = 1,$ \textit{l}}}
         \State \textit{sequence}$[i]$, \textit{labels}$[i]$ $\gets s, l_s$ $\sim$  \textit{generator}$()$ 
      \EndFor
      \\
        \Comment{Use the label of the previous instance as an 0-based \textit{address} and}
        
        \Comment{append the instance located in that \textit{address} in the sequence}
      \For{$j =  1,$ \textit{k}} 
     	\State append(\textit{sequence}, \textit{sequence}$[\textit{labels}$[$\textit{l+j-1}$]$ ]$ ) 
      \EndFor

      \State \textbf{return} \textit{sequence}\Comment{Return full sequence}
    \EndProcedure
}
  \end{algorithmic}
\end{algorithm}

\begin{algorithm}[H]
  \caption{Training Sequence Generation for Similarity-Cued Recall Tasks}\label{pseudoSimilarityDependency}
  \begin{algorithmic}[1]
\scriptsize{  
    \Procedure{SimilarityBasedDependency}{}
      \State $\textit{generator} \gets \text{instance generator from dataset (MNIST)}$
	  \State $\textit{l} \gets \text{length of random sequence}$
      \State $\textit{k} \gets \text{length of recall interval}$
      \State $\textit{sequence} \gets  \{\}$
      \\
      \Comment{Pick a random sequence of instances from the dataset}
      \For{$i = 1,$ \textit{l}}
         \State \textit{sequence}$[i]$ $\gets s$ $\sim$  \textit{generator}$()$ 
      \EndFor
      \\
\Comment{Uniformly choose a random \emph{sub}-sequence and}

\Comment{append it to the end of the sequence}

      \State $\textit{r} \sim$  Uniform[$1, l-k$]
      \State $\textit{sub-sequence} \gets  \textit{sequence}[r:r+k]$
     	\State append(\textit{sequence}, \textit{sub-sequence})

    \textbf{return} \textit{sequence}\Comment{Return full sequence}
    \EndProcedure
}
  \end{algorithmic}
\end{algorithm}

\end{document}